\documentclass[acmsmall]{acmart}
\usepackage{multirow}
\usepackage{tabularx}
\usepackage{longtable}
\usepackage{tikz}
\usepackage{pgfplots}
\usepackage{CJK}
\pgfplotsset{compat=1.17}
\usepgfplotslibrary{polar}
\usepackage{hyperref}
\usepackage{xcolor}

\AtBeginDocument{%
  \providecommand\BibTeX{{%
    \normalfont B\kern-0.5em{\scshape i\kern-0.25em b}\kern-0.8em\TeX}}}

\setcopyright{acmlicensed}
\copyrightyear{2018}
\acmYear{2018}
\acmDOI{XXXXXXX.XXXXXXX}

\acmJournal{JACM}
\acmVolume{37}
\acmNumber{4}
\acmArticle{111}
\acmMonth{8}

\begin{document}

\title{CRUD-RAG: A Comprehensive Chinese Benchmark for Retrieval-Augmented Generation of Large Language Models}

%






\author{Yuanjie Lyu}
\authornote{Both authors contributed equally to this research.}
\affiliation{%
  \institution{University of Science and Technology of China}
  \city{Hefei}
  \country{China}}
\email{s1583050085@gmail.com}

\author{Zhiyu Li}
\authornotemark[1]
\affiliation{%
  \institution{Institute for Advanced Algorithms Research (Shanghai)}
  \country{China}}
\email{lizy@iaar.ac.cn}

\author{Simin Niu}
\email{niusimin@ruc.edu.cn}
\affiliation{%
  \institution{Renmin University of China}
  \city{Beijing}
  \country{China}}

\author{Feiyu Xiong}
\email{xiongfy@iaar.ac.cn}
\author{Bo Tang}
\email{tangb@iaar.ac.cn}
\affiliation{%
  \institution{Institute for Advanced Algorithms Research (Shanghai)}
  \country{China}}

\author{Wenjin Wang}
\email{wangwj@iaar.ac.cn}
\author{Hao Wu}
\email{wuh@iaar.ac.cn}
\affiliation{%
  \institution{Institute for Advanced Algorithms Research (Shanghai)}
  \country{China}}

\author{Huanyong Liu}
\email{liuhuanyong@360.cn}
\affiliation{%
  \institution{360 AI Research Institute}
  \city{Beijing}
  \country{China}}

\author{Tong Xu}
\authornote{Corresponding author.}
\affiliation{%
  \institution{University of Science and Technology of China}
  \city{Hefei}
  \country{China}}
\email{tongxu@ustc.edu.cn}

\author{Enhong Chen}
\affiliation{%
  \institution{University of Science and Technology of China}
  \city{Hefei}
  \country{China}}
\email{cheneh@ustc.edu.cn}



\renewcommand{\shortauthors}{Lyu, et al.}

\begin{abstract}
Retrieval-Augmented Generation (RAG) is a technique that enhances the capabilities of large language models (LLMs) by incorporating external knowledge sources. This method addresses common LLM limitations, including outdated information and the tendency to produce inaccurate "hallucinated" content. However, evaluating RAG systems is a challenge. Most benchmarks focus primarily on question answering applications, neglecting other potential scenarios where RAG could be beneficial. Accordingly, in the experiments, these benchmarks often assess only the LLM components of the RAG pipeline or the retriever in knowledge-intensive scenarios, overlooking the impact of external knowledge base construction and the retrieval component on the entire RAG pipeline in non-knowledge-intensive scenarios. To address these issues, this paper constructs a large-scale and more comprehensive benchmark, and evaluates all the components of RAG systems in various RAG application scenarios. Specifically, we refer to the CRUD actions that describe interactions between users and knowledge bases, and also categorize the range of RAG applications into four distinct types—Create, Read, Update, and Delete (CRUD). "Create" refers to scenarios requiring the generation of original, varied content. "Read" involves responding to intricate questions in knowledge-intensive situations. "Update" focuses on revising and rectifying inaccuracies or inconsistencies in pre-existing texts. "Delete" pertains to the task of summarizing extensive texts into more concise forms. For each of these CRUD categories, we have developed different datasets to evaluate the performance of RAG systems. We also analyze the effects of various components of the RAG system, such as the retriever, context length, knowledge base construction, and LLM. Finally, we provide useful insights for optimizing the RAG technology for different scenarios\footnote{The source code is available at GitHub: \url{https://github.com/IAAR-Shanghai/CRUD_RAG}}.

\end{abstract}



\received{20 February 2007}
\received[revised]{12 March 2009}
\received[accepted]{5 June 2009}

\maketitle
\begin{figure}[t]
\centering
\includegraphics[width=\textwidth]{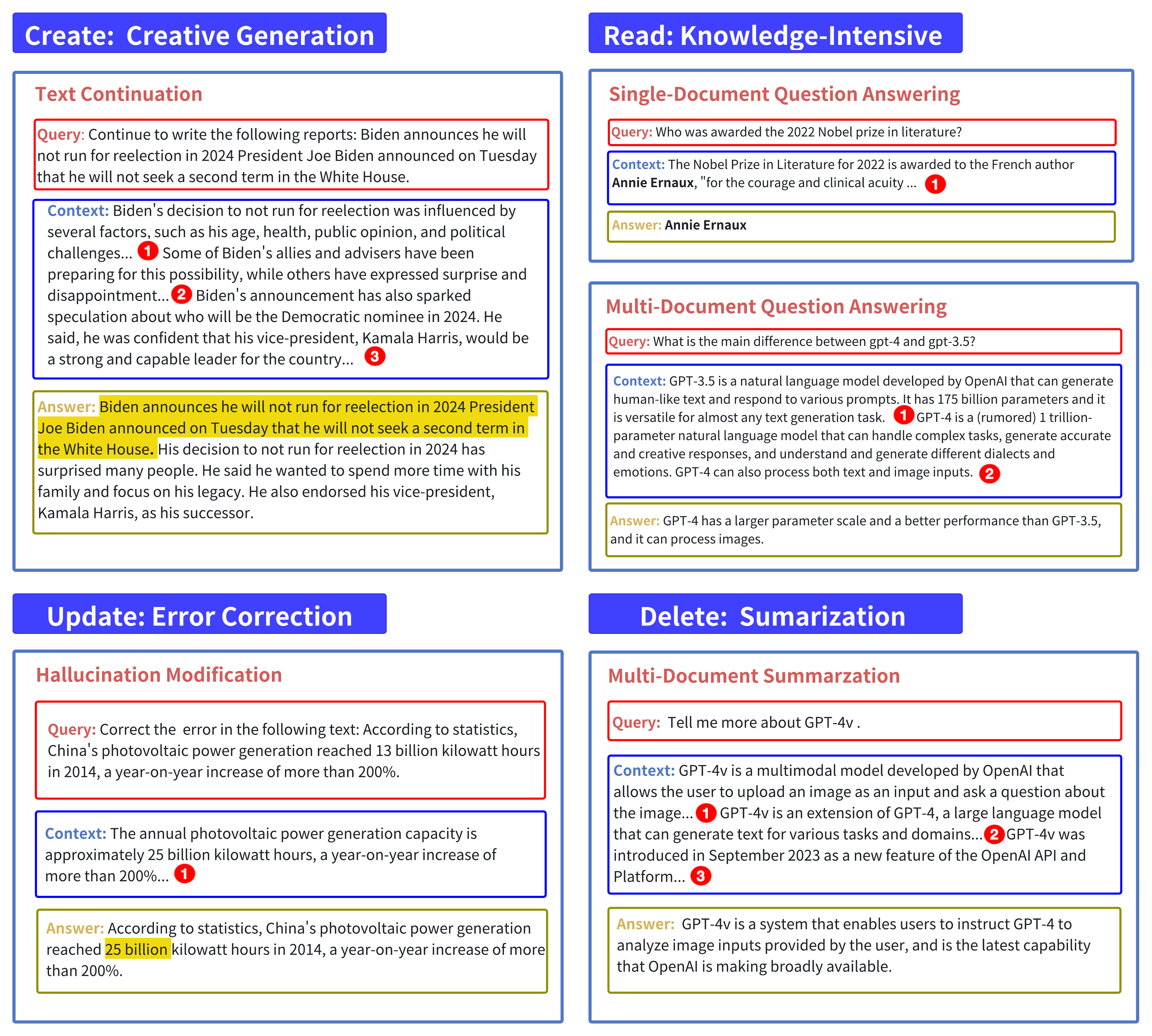}
\caption{
We have classified the application scenarios of RAG into four primary aspects: Create, Read, Update, and Delete. The figure provides an illustrative example for each category, showcasing the wide-ranging potential of RAG technology.}
\label{fig:intro}
\end{figure}

\section{Introduction}
Retrieval-augmented generation (RAG) is an advanced technique that leverages external knowledge sources to enhance the text generation capabilities of large language models(LLMs). It retrieves relevant paragraphs from a corpus based on the input, and feeds them to the LLMs along with the input. With the help of external knowledge, LLMs can generate more accurate and credible responses and effectively address challenges such as outdated knowledge~\cite{he2022rethinking}, hallucinations~\cite{lyu2022faithful,zuccon2023chatgpt,benedict2023gen,chen2023hallucination}, and lack of domain expertise~\cite{li2023chatgpt,shen2023chatgpt}. Therefore, RAG technology is attracting increasing attention.

Although the effectiveness of retrieval-augmented strategies has been proven through extensive practice, their implementation still requires a significant amount of tuning. The overall performance of the RAG system is affected by multiple factors, such as the retrieval model, construction of the external knowledge base, and language model. Therefore, automatic evaluation of RAG systems is crucial. Currently, there are only a few existing benchmarks for evaluating RAG performance, as creating high-quality datasets and experimenting with them entail significant costs. These benchmarks can be classified into two types: reference-required and reference-free evaluation. 
Reference-free evaluation frameworks, such as RAGAS~\cite{es2023ragas} and ARES~\cite{saad2023ares}, use LLM-generated data to evaluate RAG systems on contextual relevance, faithfulness, and informativeness. These frameworks do not depend on ground truth references, but only assess the coherence of the generated text with the retrieved context. This approach may be unreliable if the retrieved external information is low-quality.

Consequently, reference-required evaluations remain the predominant method for assessing RAG systems. Existing benchmarks for reference-required evaluations, such as RGB~\cite{chen2023benchmarking} and NQ~\cite{kwiatkowski2019natural}. do have their limitations. First, they all rely on question answering tasks to measure the performance of RAG systems. Question answering is not the only RAG application scenario, and an optimization strategy that works well for question answering may not be generalized to other scenarios. Thus, these benchmarks may not capture the full potential of RAG systems. Second, in the experiments, current evaluations usually focus on evaluating the LLM part of the RAG pipeline, or focus on retriever performance in the knowledge-intensive scenario~\cite{petroni2020kilt},  while ignoring the retrieval methods in non-knowledge-intensive scenarios and external knowledge base construction. These components are also crucial for RAG systems.
Therefore, a comprehensive evaluation of the RAG system may not be obtained using any existing benchmarks.

To evaluate the performance of RAG in different application scenarios, we need a comprehensive benchmark that covers more than just the question-answering task. ~\citet{DBLP:conf/nips/LewisPPPKGKLYR020} argue that the core of RAG systems is their interactive way of combining LLMs with external knowledge sources. And following ~\cite{kilov1990semantic}, we can group any interaction with external knowledge sources into four basic actions: create, read, update, and delete, which are also known as CRUD actions~\cite{truica2015performance}. Therefore, we can use the CRUD framework to classify the RAG systems' application scenarios. As shown in Figure \ref{fig:intro}, each CRUD category demonstrates different capabilities of the RAG system: 

\begin{itemize}
\item In "\textbf{CREATE}", the system improves the input text by adding relevant information from external sources, making creative outputs such as poetry, stories, or code. 
\item In "\textbf{READ}", the system uses external knowledge retrieval to answer questions, solve problems in question-answering, dialogue, and reasoning, and increase understanding of the input text. 
\item In "\textbf{UPDATE}", the system fixes errors in the input text using retrieved content, correcting spelling, grammar, or factual errors to make the text better. 
\item In "\textbf{DELETE}", the system simplifies the input by improving retrieval results, removing unnecessary details, and doing tasks like text summarization or simplification. 
\end{itemize}

To evaluate the RAG system in these four scenarios, we introduce CRUD-RAG, a comprehensive, large-scale Chinese RAG benchmark. CRUD-RAG consists of four evaluation tasks: text continuation, question answering (with single-document and multi-document questions), hallucination modification, and open-domain multi-document summarization, which respectively correspond to the CRUD-RAG classification of RAG application scenarios. We construct CRUD-RAG by crawling the latest high-quality news data from major news websites in China, which aims to minimize the likelihood of LLMs encountering these data during training. Then, we automatically create datasets using GPT-4 based on these news data.  For the multi-document summarization task, we apply a reverse construction strategy. We first generate news events and their summaries using GPT-4. Then, we use these events as keywords to search for 10 related and non-duplicate reports from the web, which we add to our retrieval database. During evaluation, the RAG system will use the retrieval database to generate summaries for the events. For the text continuation task, we split the news text into a beginning and a continuation paragraph. We then use each sentence in the continuation paragraph as a keyword to search for 10 related reports on the Web. We remove any duplicate content and add the reports to the retrieval database. For the single-document QA task, we use the RGB~\cite{chen2023benchmarking} construction method. For the multi-document QA task, we use the Chain-of-Thought technology to help the model identify common and different aspects among documents, and then generate questions based on these aspects with increasing difficulty. For the hallucination modification task, we use the annotations in the UHGEval dataset and correct hallucinations with GPT-4. We also include the real news in UHGEval in the retrieval database.

In the experiments, we systematically evaluate the RAG system's performance on our CRUD-RAG benchmark. We also investigate various factors that affect the RAG system, such as the context length, the chunk size, the embedding model, the retrieval algorithms, and the LLM. Based on our experimental results, we provide some valuable suggestions for building effective RAG systems.

The contributions of this paper are:
\begin{itemize}

\item \textbf{A comprehensive evaluation benchmark}: Our benchmark covers not only question answering, but also create, read, update, and delete (CRUD) of RAG applications. 

\item \textbf{High-quality evaluation datasets}: We constructed diverse datasets for different evaluation tasks, based on the application scenarios of RAG. These tasks include text continuation, multi-document summarization, question answering, and hallucination modification.

\item \textbf{Extensive experiments}: we performed extensive experiments on our benchmark, using various metrics to measure the performance of RAG systems. Based on our experiments, we offered useful guidance for future researchers and RAG system developers.
\end{itemize}


\section{Related Work}
\subsection{Retrieval-Augmented Generation}
LLMs excel in text generation but also confront challenges such as outdated knowledge and the generation of hallucinatory content~\cite{he2022rethinking,DBLP:conf/emnlp/CaoDWC20,DBLP:conf/naacl/RaunakMJ21}. In response to these challenges, RAG, also referred to as RALM (Retrieval-Augmented Language Models), incorporates external knowledge to generate responses characterized by enhanced accuracy and realism~\cite{DBLP:journals/corr/abs-2301-12652}. This is particularly critical in domains that heavily depend on precision and reliability, including but not limited to the legal, medical, and financial sectors. Retrieval models have been promoting the development of language models~\cite{10.1145/3639818,zhai2004study,ganguly2015word}.

Conventional RAG systems adhere to a standardized workflow encompassing indexing, retrieval, and generation phases~\cite{DBLP:conf/nips/LewisPPPKGKLYR020, DBLP:journals/corr/abs-2305-14283}. The indexing phase encompasses data cleansing, extraction, transformation into plain text, segmentation, and indexing, utilizing embedding models to transform text fragments into vector representations~\cite{DBLP:journals/corr/abs-2312-10997,DBLP:conf/acm/AsaiMZC23}. In the retrieval phase, the system computes similarity scores based on the user's query to select the most pertinent text fragments. In the generation phase, the query and selected documents are amalgamated into prompts, facilitating the LLMs in generating a response. While this method is straightforward, it encounters challenges related to retrieval quality, generation quality, and enhancement processes~\cite{DBLP:conf/emnlp/JiangXGSLDYCN23,DBLP:journals/corr/abs-2208-03299}.

In response to these challenges, researchers concentrate on the enhancement of the retriever, a task that can be categorized into three key aspects: pre-retrieval processing, retrieval model optimization, and post-retrieval processing~\cite{ART2023}. Pre-retrieval processing encompasses data transformer to enhance text standardization, ensuring factual accuracy, optimizing index structures, adjusting block sizes, and rewriting query~\cite{DBLP:journals/corr/abs-2308-09729,AAS2023,DBLP:conf/emnlp/WangYW23,DBLP:conf/acl/GaoMLC23}. Retrieval model optimization entails the fine-tuning of domain-specific embedding models and the application of dynamic embedding techniques~\cite{DBLP:conf/iclr/DaiZMLNLBGHC23,DBLP:journals/corr/abs-2310-07554}. Post-retrieval processing minimizes context length through reranking and compression operations, aiming to emphasize critical information, diminish noise interference, and enhance integration and utilization by the generator~\cite{DBLP:conf/emnlp/YangLZW0L023,DBLP:journals/corr/abs-2310-04408,DBLP:conf/emnlp/Ma0HS23}.

Furthermore, to enhance the precision and efficiency of the generator when handling retrieval content, scholars have undertaken a series of optimization measures. As an illustration, researchers have devised methods such as Chain-of-Note (CON) for the generator~\cite{DBLP:journals/corr/abs-2311-09210}. CON generates continuous reading notes to comprehensively evaluate the relevance of retrieved documents to the posed question, integrating this information to produce precise final responses. This approach further enhances the capability of RAG in managing retrieval information, guaranteeing the production of responses that are simultaneously accurate and pertinent. In specific domains, such as medical and legal, models undergo fine-tuning to enhance the generator's performance within those particular fields~\cite{DBLP:journals/corr/abs-2305-02437,DBLP:journals/corr/abs-2305-18846,DBLP:conf/emnlp/YeLDLLSL20}. Through the implementation of these methods, the generator can more effectively process retrieved information and furnish responses that are more accurate and relevant.

\subsection{RAG Benchmarks}

\begin{table}[t]
\centering
\caption{Relate Work.}
\label{tab:rw}

\resizebox{\textwidth}{!}{
\begin{tabular}{@{}cccccccc@{}}
\toprule[1pt]
\textbf{Method}                                        & \textbf{Dataset}  
                           & \textbf{Scale}  

& \textbf{Evaluation Metrics}                                                                                                                              & \textbf{Evaluation Method}                                                                                 & \textbf{Application Field}                                                     & \textbf{Ref.}& \textbf{Lang.} \\ \midrule
~\cite{ERABT2023}        & \begin{tabular}[c]{@{}c@{}}Based on LangChain Python\\ documentation QA dataset\end{tabular} 
& 86
& \begin{tabular}[c]{@{}c@{}}Accuracy of answer, Faithfulness of \\ response to the retrieved document\end{tabular}                                        & \begin{tabular}[c]{@{}c@{}}Evaluating retrieval and\\  generation consistency\end{tabular}                 &\begin{tabular}[c]{@{}c@{}} General QA scenarios\\ (\textbf{R}ead)\end{tabular}                                                         & Yes      & EN          \\ \midrule
~\cite{ERABT2023}   & \begin{tabular}[c]{@{}c@{}}PDF documents containing\\ tables and charts\end{tabular}            
& 5
& \begin{tabular}[c]{@{}c@{}}Accuracy of answer, Faithfulness of \\ response to the retrieved document\end{tabular}                                        & \begin{tabular}[c]{@{}c@{}}Evaluating retrieval and\\  generation consistency\end{tabular}                 & \begin{tabular}[c]{@{}c@{}}Semi-structured \\ data scenarios\\(\textbf{R}ead)\end{tabular}      & Yes           & EN             \\\midrule

~\cite{DBLP:conf/emnlp/LiuZL23}                                                                & \begin{tabular}[c]{@{}c@{}}Query and responses \\ (with citations)\end{tabular}   & 1450 
& \begin{tabular}[c]{@{}c@{}}Fluency, Perceived utility, \\ Citation recall and precision\end{tabular}                                                     & Human evaluation                                                                                           & \begin{tabular}[c]{@{}c@{}} Citation\\ (\textbf{R}ead)\end{tabular}                                                                        & Yes         & EN               \\\midrule
   ~\cite{DBLP:conf/emnlp/GaoYYC23,DBLP:conf/aaai/JiLDN24}                                                                 & \begin{tabular}[c]{@{}c@{}}Questions, answers\\ and contexts\\ (with citations)\end{tabular}                                         & -                                       & \begin{tabular}[c]{@{}c@{}}Fluency, Correctness, \\  Citation quality\end{tabular}                                                      & \begin{tabular}[c]{@{}c@{}}Self-devised metrics,\\Human evaluation \end{tabular} & \begin{tabular}[c]{@{}c@{}} Citation\\ (\textbf{R}ead)\end{tabular}                                                                          & Yes                & EN

   \\\midrule
      ~\cite{xu2024aliiceevaluatingpositionalfinegrained}                                                                 & \begin{tabular}[c]{@{}c@{}}Questions, answers\\ and contexts\\ (with citations)\end{tabular}                                         & 1948                                       & \begin{tabular}[c]{@{}c@{}}Location citation recall, Location precision,\\The coefficient of variation of citation locations\end{tabular}                                                      & \begin{tabular}[c]{@{}c@{}}Self-devised metrics\end{tabular} & \begin{tabular}[c]{@{}c@{}} Citation\\ (\textbf{R}ead)\end{tabular}                                                                          & Yes                & EN

   \\\midrule

         ~\cite{li2024citationenhancedgenerationllmbasedchatbots}                                                                 & \begin{tabular}[c]{@{}c@{}}Questions, answers\\ and contexts\\ (with citations)\end{tabular}                                         & 3422                                       & \begin{tabular}[c]{@{}c@{}}Accuracy of factual and non-factual, \\AUC-PR, and so on\end{tabular}                                                      & \begin{tabular}[c]{@{}c@{}}Self-devised metrics, \\Common metrics\end{tabular} & \begin{tabular}[c]{@{}c@{}} Citation\\ (\textbf{R}ead)\end{tabular}                                                                          & Yes                & EN

   \\\midrule
       ~\cite{zhang2024finegrainedcitationevaluationgenerated}                                                                 & \begin{tabular}[c]{@{}c@{}}Statement-citation Pairs\end{tabular}                                         & 12681                                       & \begin{tabular}[c]{@{}c@{}}Correlation, Classification performance,\\Retrieval effectiveness, Faithfulness\end{tabular}                                                      & \begin{tabular}[c]{@{}c@{}}Self-devised metrics,\\Common metrics,\\Human evaluation\end{tabular} & \begin{tabular}[c]{@{}c@{}} Citation\\ (\textbf{R}ead)\end{tabular}                                                                          & Yes                & EN

   \\\midrule
   ~\cite{cao2024verifiablegenerationsubsentencelevelfinegrained}                                                                 & \begin{tabular}[c]{@{}c@{}} Paragraphs\\ (with citations)\end{tabular}                                         & 10000                                       & \begin{tabular}[c]{@{}c@{}}Citation density,\\The coverage of reference facts\end{tabular}                                                      & \begin{tabular}[c]{@{}c@{}}Self-devised metrics\end{tabular} & \begin{tabular}[c]{@{}c@{}} Citation\\ (\textbf{R}ead)\end{tabular}                                                                          & Yes                & EN

   \\\midrule
~\cite{HELR2023} & \begin{tabular}[c]{@{}c@{}}Questions, answers\\ and contexts \end{tabular}
& 200
& \begin{tabular}[c]{@{}c@{}}Categorization ability, \\ Logical/Mathematical reasoning, \\ Complex question solving, \\ Summarization ability\end{tabular} & Accuracy                                                                                                   & \begin{tabular}[c]{@{}c@{}}Financial services, \\ Legal, Business\\(\textbf{R}ead, \textbf{D}elete)\end{tabular} & Yes      & EN                  \\\midrule
~\cite{chen2023benchmarking}                                                                  & LLM-generated dataset                          & 1000                                                 & \begin{tabular}[c]{@{}c@{}}Noise robustness, \\ Negative rejection, \\ Information integration, \\ Counterfactual robustness\end{tabular}                & Self-devised metrics                                                                                       & \begin{tabular}[c]{@{}c@{}}General, especially \\ news domain\\(\textbf{R}ead, \textbf{U}pdate)\end{tabular}     & Yes              &  \begin{tabular}[c]{@{}c@{}}CN\\  EN\end{tabular}           \\\midrule
~\cite{TruLens-Eval2023}                                                         & —                                                      & —                                              & \begin{tabular}[c]{@{}c@{}}Context relevance, Groundedness, \\ Answer relevance\end{tabular}                                                             & Analyzing the RAG triad                                                                                    &              \begin{tabular}[c]{@{}c@{}} General\\ (\textbf{C}reate, \textbf{R}ead)\end{tabular}                                                            & No            & —             \\\midrule
~\cite{es2023ragas}                                                                & —                                                            & —                                       & \begin{tabular}[c]{@{}c@{}}Faithfulness, Answer relevance, \\ Context relevance\end{tabular}                                                             & \begin{tabular}[c]{@{}c@{}}Automated evaluation\\ using LLM prompts\end{tabular}                           & \begin{tabular}[c]{@{}c@{}} General\\ (\textbf{C}reate, \textbf{R}ead)\end{tabular}                                                                          & No            & —             \\\midrule
~\cite{saad2023ares}                                                                 & LLM-generated dataset                                         & 150                                       & \begin{tabular}[c]{@{}c@{}}Context relevance, Answer faithfulness, \\ Answer relevance\end{tabular}                                                      & \begin{tabular}[c]{@{}c@{}}Generating custom LLM judges for \\ each component of a RAG system\end{tabular} & \begin{tabular}[c]{@{}c@{}} General\\ (\textbf{C}reate, \textbf{R}ead)\end{tabular}                                                                          & No                & EN

   \\\midrule

\textbf{Ours}                                                                 & LLM-generated dataset                                         & 36166                                       & \begin{tabular}[c]{@{}c@{}} ROUGE, BLEU,\\ bertScore, RAGQuestEval\end{tabular}                                                      & \begin{tabular}[c]{@{}c@{}}Evaluating retrieval and\\  generation consistency\end{tabular} & \begin{tabular}[c]{@{}c@{}} General\\ (\textbf{C}reate, \textbf{R}ead\\\textbf{U}pdate, \textbf{D}elete)\end{tabular}                                                                          & Yes                & CN      
\\ \bottomrule[1pt]
\end{tabular}
}
\end{table}
When investigating the development and optimization of RAG, the effective evaluation of their performance becomes a fundamental concern. Table \ref{tab:rw} shows some commonly used benchmarks for evaluating RAG. LangChain provides benchmark tasks, such as LangChain Docs Q\&A and Semi-structured Reports~\cite{ERABT2023}, designed to assess various RAG architectures. These datasets are constructed from snapshots of Python documentation and PDFs containing tables and charts. They emphasize the model's capability to handle structured and semi-structured data. Evaluation standards encompass the accuracy of answers and the faithfulness of model responses. Utilizing large models for question-answering generation has emerged as a prevalent approach in building evaluation datasets. For instance, RGB~\cite{chen2023benchmarking} creates its evaluation dataset by gathering recent news reports and employing LLM to generate relevant events, questions, and answers. Conversely, ARES~\cite{saad2023ares}. relies on generating synthetic queries and answers, leveraging the FLAN-T5 XXL model. These methods not only showcase the RAG system's proficiency in handling real-time data but also illustrate the utility of automation and synthetic data in the evaluation process. For evaluating the capabilities of models across various professional domains, the Instruct-Benchmark-Tester dataset encompasses a range of question types, with a particular focus on financial services, legal, and intricate business scenarios~\cite{HELR2023}.

Depending on whether the evaluation phase incorporates ground truth, metrics of existing evaluation methods can be categorized into those necessitating reference and those not requiring it. Reference-required evaluation methods gauge the accuracy and robustness of the RAG by contrasting model-generated answers with factual benchmarks. As an example, RAG-Instruct-Benchmark-Tester~\cite{HELR2023} employs accuracy score as an evaluation metric, a widely acknowledged measure of model performance that assesses the extent to which model-generated answers align with reference answers. The primary objective of RGB~\cite{chen2023benchmarking} is to evaluate whether large models can effectively utilize external documents to acquire knowledge and generate accurate answers. Its evaluation metrics encompass accuracy, rejection rate, error detection rate, and correction rate.

Reference-free evaluation methods, including TruLens-Eval~\cite{TruLens-Eval2023}, RAGAS~\cite{es2023ragas}, and ARES~\cite{saad2023ares}, provide distinct viewpoints for evaluating the performance of RAG systems, particularly concerning context relevance, answer faithfulness, and answer relevance. TruLens-Eval~\cite{TruLens-Eval2023} introduces the RAG Triad as an innovative approach to evaluate hallucination issues within the RAG architecture, encompassing context relevance, groundedness, and answer relevance. RAGAS~\cite{es2023ragas}, serving as a reference-free evaluation framework, concentrates on assessing the retrieval system's capacity to identify pertinent and concentrated context passages, along with the LLMs' proficiency in faithfully and accurately leveraging these passages. In contrast to RAGAS, which depends on a predefined set of heuristically crafted prompts, ARES generates tailored LLMs judges for each aspect of a RAG pipeline, leading to a substantial enhancement in evaluation precision and accuracy when compared to existing methods such as RAGAS. Furthermore, ARES~\cite{saad2023ares} employs prediction-powered inference to offer statistical assurances for its scoring, generating confidence intervals. ARES emphasizes three evaluation scores: context relevance, answer faithfulness, and answer relevance, highlighting the importance of a proficient RAG system in identifying relevant contexts and producing both faithful and relevant answers.
Regarding evaluation methods, ~\cite{DBLP:conf/emnlp/LiuZL23} places an emphasis on assessing the credibility and accuracy of responses generated by generative search engines through manual inspection. Nonetheless, manual evaluation possesses drawbacks, including high costs and challenges in scalability. Hence, rule-based evaluation metrics such as accuracy, exact match, rouge, or self-devised metrics like rejection rate, error detection rate, and correction rate continue to be widely adopted in the field. Furthermore, employing LLMs for evaluation closely approximates manual evaluation outcomes.

\subsection{Citation-Enhanced RAG}
In traditional RAG methods, despite the rich information sources provided by retrieved contexts for text generation, these models often do not explicitly require responses to provide corresponding citations, making traceability difficult. 
Therefore, enhancing text verifiability by introducing citation links, i.e., explicit references, has become an important research direction in the RAG field~\cite{DBLP:conf/emnlp/GaoYYC23,xu2024aliiceevaluatingpositionalfinegrained}.

Providing citation indicators in the response text offers several clear benefits. First, users can easily verify the claims made by LLMs based on the provided citations, thus improving the transparency and credibility of the text. Second, if the text generated by LLMs adheres faithfully to the cited contexts, it can significantly improve its accuracy and reduce the phenomenon of "hallucinations"~\cite{DBLP:conf/emnlp/GaoYYC23}. Given this, generating high-quality citations and evaluating the quality of citation generation have become crucial elements of assessing RAG performance. Constructing appropriate prompts directly through the retrieval context to guide the model in generating corresponding citations constitutes a direct and effective method of citation generation~\cite{DBLP:conf/aaai/JiLDN24}.

In terms of evaluation, early research primarily focused on the fluency, accuracy, and basic citation quality of the text generated by LLMs~\cite{DBLP:conf/emnlp/GaoYYC23, li2024citationenhancedgenerationllmbasedchatbots}. For example, Rashkin et al. proposed the "Attributable to Identified Sources" (AIS) score~\cite{10.1162/coli_a_00486}, which serves as a valuable tool for measuring the degree to which generated text is faithful to its sources. As research progressed, scholars recognized the need for more detailed evaluation methods to differentiate between various levels of citation support. By creating specialized datasets such as SCIFI~\cite{cao2024verifiablegenerationsubsentencelevelfinegrained}, researchers can more precisely evaluate fine-grained citations at the clause level in texts generated by LLMs. The ALiiCE framework~\cite{xu2024aliiceevaluatingpositionalfinegrained}, by analyzing the atomic structure of sentence claims, introduced fine-grained evaluation metrics, such as location citation recall and precision, and the coefficient of variation of citation locations, to more granularly evaluate the quality of citation generation in RAG~\cite{xu2024aliiceevaluatingpositionalfinegrained}. In practical applications, ~\cite{zhang2024finegrainedcitationevaluationgenerated} found that RAG requires more complex evaluation frameworks to distinguish between various levels of citation support by comparing different fidelity metrics. These RAG evaluation methods not only consider the presence of citations but also their accuracy and relevance. 

While Citation-Enhanced RAG delves deeply into the specific domain of citation generation, aiming to improve the credibility and accuracy of text generated by RAG systems, our benchmark provides a comprehensive evaluation framework encompassing various aspects of RAG systems and multiple application scenarios.

\begin{figure}[t]
\centering
\includegraphics[width=\textwidth]{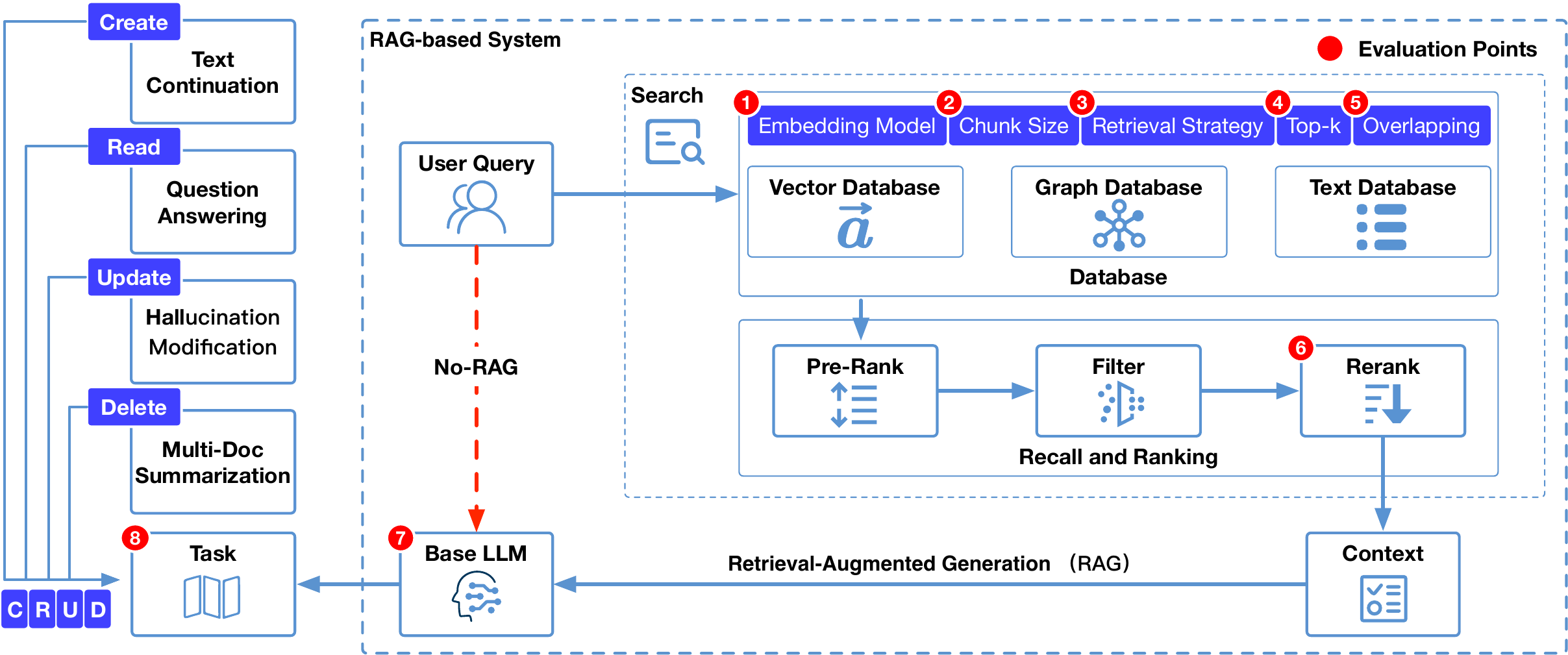}
\caption{
Illustration of CRUD-RAG, our comprehensive Chinese benchmark for RAG. It classifies the RAG application scenarios into four categories: create, read, update, and delete. For each category, we create appropriate evaluation tasks and datasets. In the experiments, we evaluate various components of the RAG system using our benchmarks.
}
\label{fig:benchmark}
\end{figure}

\section{CRUD-RAG: A Comprehensive Chinese Benchmark for RAG}

As we discussed earlier, implementing RAG effectively requires careful tuning of multiple components, such as the retrieval model, the knowledge corpus, the language model, and the query formulation. Therefore, we need a framework that can evaluate the RAG system automatically. This framework would enable us to examine how these components affect the system's performance, and provide us with useful insights for improving and innovating the system.

However, The current RAG benchmarks have several drawbacks: 
they only evaluate question answering tasks~\cite{qu2020open,aliannejadi2019asking,yu2020few}, ignoring other diverse application of RAG. The optimization strategy for question-answering tasks may not suit other tasks; And in the evaluation experiment, current RAG benchmarks only account for the LLM component in the RAG pipeline, or the retriever in the knowledge-intensive scenario, disregarding the vital roles of retrieval database construction and retrieval in non-knowledge-intensive scenarios.


To address the shortcomings of previous benchmarks, we introduce CRUD-RAG, a comprehensive Chinese benchmark for RAG. Figure \ref{fig:benchmark} illustrates the features of our CRUD-RAG benchmark. It classifies the RAG application scenarios into four categories: Create, Read, Update, and Delete, then we construct appropriate evaluation tasks and datasets for each category. Besides, in the experiments, we will assess the impact of various components of RAG, such as chunk size, retrieval strategy, top-k, LLM, etc., on all tasks.

\begin{figure}[t]
\centering
\includegraphics[width=\textwidth]{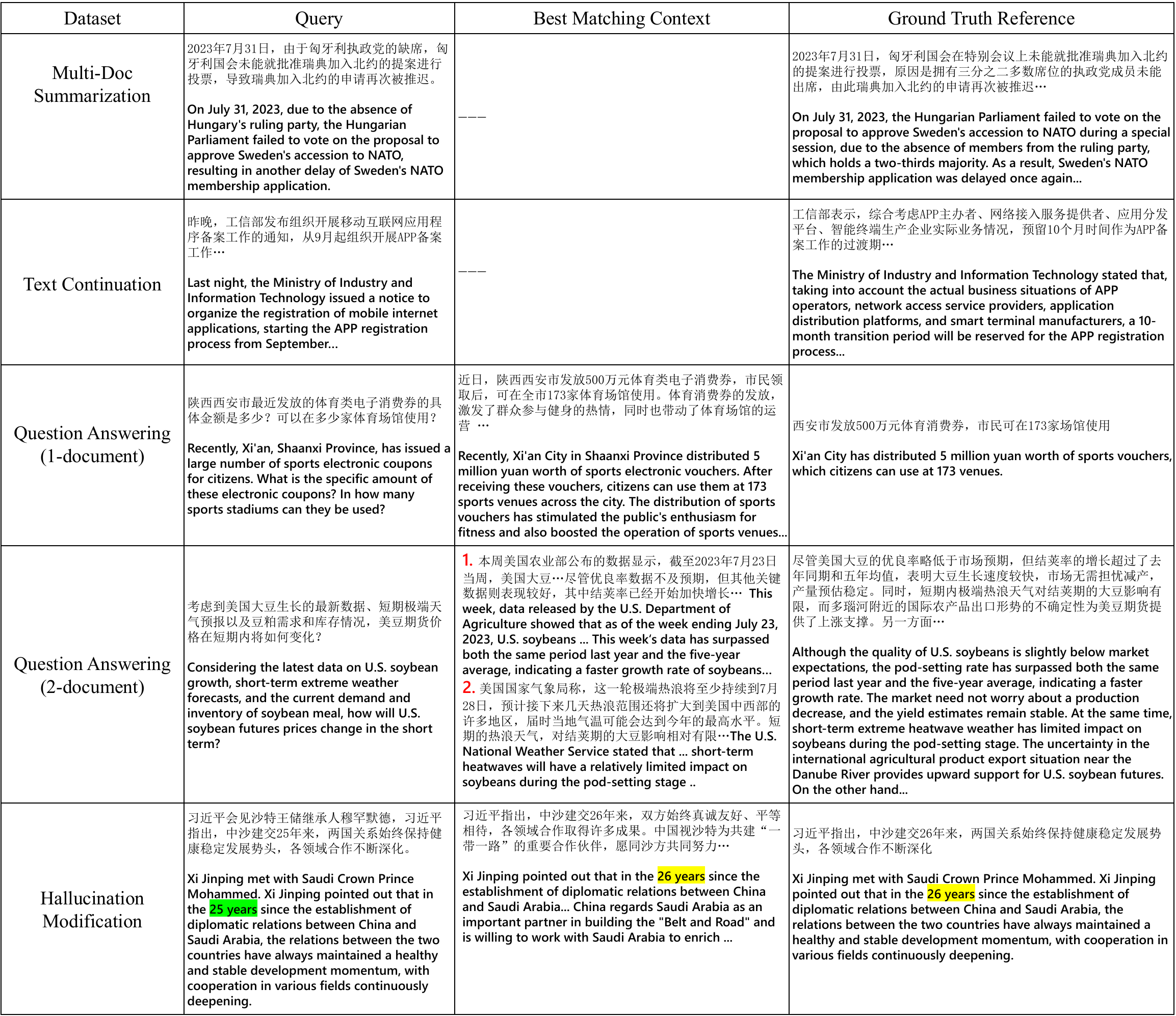}
\caption{Some examples of the datasets we constructed. We did not provide the best matching context for the multi-document summarization dataset and the text continuation dataset, because these two datasets were built in a reverse way, and the context matching degree for these two tasks was rather vague.}
\label{fig:example}
\end{figure}

In the following section, we will describe the evaluation tasks and the datasets that we design for each RAG application scenario type.  We select text continuation, question answering (single and multi-document), hallucination modification, and multi-document summarization as representative tasks in the CRUD (Create, Read, Update, Delete) scenario and construct corresponding datasets. The summarization (D) and continuation (C) datasets were constructed simultaneously, since the construction of both datasets requires the use of a search engine.  They will be discussed together in the following section. The construction of the question-answering(R) and hallucination modification(D) datasets is relatively independent. To maintain narrative coherence, we will introduce the dataset construction process in the order of DCRU. Table \ref{tab:dataset} presents the size and composition of our datasets, and Figure \ref{fig:example} illustrates an example of our datasets.

\begin{table}[t]
\centering
\caption{The composition of our datasets.}
\label{tab:dataset}

\resizebox{\textwidth}{!}{
\begin{tabular}{m{3.5cm}m{2cm}m{5cm}m{6cm}}
\toprule[1pt]
Dataset Name & Dataset Size & Components & Evaluation objectives  \\
\midrule
Text Continuation & 10,728 & An initial part of an article, followed by its extension or completion. & Evaluate the RAG system's performance in "Create" scenarios(Creative generation). \\
\midrule
Question Answering\newline(1-document) & 3,199 & A collection of question-answer pairs, where the answer is directly extractable from a document passage. & Evaluate the RAG system's performance in "Read" scenarios(Knowledge-intensive application). \\
\midrule
Question Answering\newline(2-document) & 3,192 & A collection of question-answer pairs, where the answer \textbf{requires synthesis of information from 2 different document sources}. & The objective is the same as 1-document QA, but it also examines \textbf{the reasoning ability of combining 2 documents}. \\
\midrule
Question Answering\newline(3-document) & 3,189 & A collection of question-answer pairs, where the answer \textbf{requires synthesis of information from 3 different document sources}. & The objective is the same as 1-document QA, but it also examines \textbf{the reasoning ability of combining 3 documents}.\\
\midrule
Hallucination\newline Modification & 5,130 &  Some sentences containing errors, and the sentence with the errors fixed. & Evaluate the RAG system's performance in "Update" scenarios(Error correction application). \\
\midrule
Multi-Doc \newline Summarization & 10,728 & A one-sentence headline of an article, followed by a brief summary of the article. & Evaluate the RAG system's performance in "Delete" scenarios(Summarization). \\
 \\
 \midrule
Retrieval Database & 86,834 & As the knowledge base for the RAG system, we expect the RAG system to retrieve relevant content from the knowledge base to address the above tasks. & --------- \\
\bottomrule[1pt]
\end{tabular}
}
\end{table}

\subsection{News Collection} 

As mentioned above, the existing benchmarks for evaluating RAG systems are mainly constructed for question answering tasks. Therefore, the datasets, such as NQ~\cite{kwiatkowski2019natural} and RGB~\cite{chen2023benchmarking}, are also tailored for this type of task. Hence, we need to construct new datasets.

We argue that the latest news data is the most suitable choice for creating an RAG evaluation dataset. Unlike other types of data, such as encyclopedias, questions, or conversations, the latest news minimizes the possibility that the model has been exposed to similar content during training. This dependency on external retrieval mechanisms allows for a comprehensive evaluation of the entire RAG process, not just the model's generation ability.  Additionally, news data is easy to collect, enabling us to maintain dataset timeliness. When the existing dataset loses its timeliness, we can quickly gather the latest news to rebuild a more challenging dataset. Moreover, the latest news data offer rich and diverse topics and content, which can test the model's performance and adaptability in various domains and situations.


Therefore, we select news as the base of our datasets. To ensure the authenticity and currency of the datasets, we collected nearly $300,000$ of historical news articles from major Chinese news websites published after July 2023, which were not exposed to the LLMs during the training phase. We remove duplicate news documents from $300,000$ news articles and filter out those that are too long or too short. We ended up with more than $80,000$ news articles. Based on the news corpus we collected, we constructed our datasets for three tasks, namely open-domain multi-document summarization, text continuation, and question answering.

\subsection{Open-domain Multi-document Summarization:
RAG Application in "Delete"}

In one of the RAG's application scenarios, "Delete", the RAG system retrieves key information from external sources based on the input text, and eliminates redundancy and irrelevance, to generate concise summaries. A suitable task for evaluating this scenario is multi-document summarization, which aims to generate a brief and coherent summary from a set of related documents. For the news data we collect, this task involves retrieving major media reports on a news event, and summarizing the background, process, and results of the event.

However, constructing such a dataset is extremely challenging. First, news articles retrieved based on events may not be fully relevant, requiring manual filtration to identify the correct and pertinent documents. Then, when generating summaries from these documents, it is essential to eliminate a significant amount of redundant information, retaining only the most important content. These tasks require manual annotation, which consumes substantial time and financial resources, and often results in too much redundant information.

\begin{figure}[t]
\centering
\includegraphics[width=\textwidth]{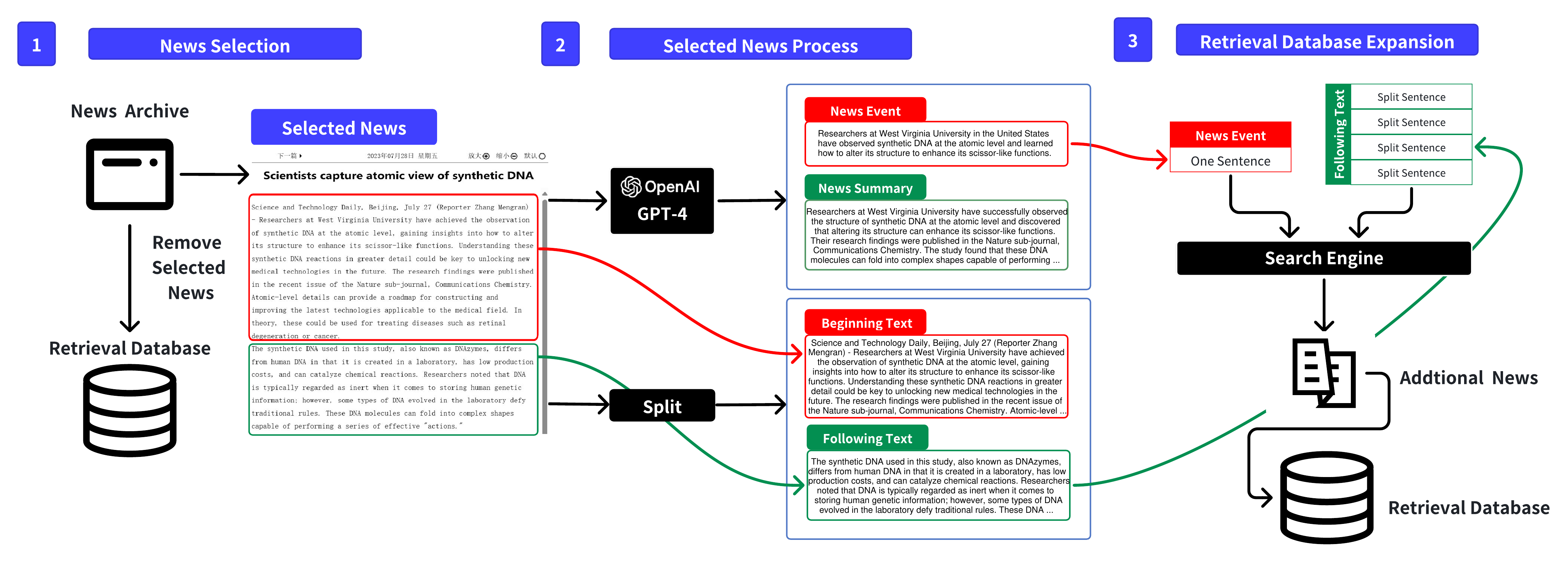}
\caption{
The dataset construction pipeline for text continuation and multi-document summarization task.
}
\label{fig:add_delete}
\end{figure}

Fortunately, we can use an existing method, which constructs a multi-document summary dataset in reverse~\cite{liu2019hierarchical}. Figure \ref{fig:add_delete} shows the construction process of multi-document summarization. In particular, our dataset construction process is as follows:
\begin{itemize}
    \item Instead of generating event summaries based on multiple related news content, we first acquire a news article from a high-quality corpus, and annotate its summary and events.
    \item Then, we search for external reference materials related to the current news by using the event text, ensuring they are connected but not the same. We conduct extensive searches to gather sufficient information to reconstruct the summary of the selected news. 
    \item In this manner, the reference literature we collect, along with the summary of the current news, collectively form a dataset of multi-document summarization.
\end{itemize}

Specifically, we first select $10,000$ news articles $d$ from our high-quality news corpus $D$, and then use GPT-4 to generate summaries and events for each article. Next, we use the events as keywords, and search for the most relevant 10 news articles on Baidu, excluding any data that is too similar to the original article. We repeat this process for all the articles, and add the expanded articles to our news corpus, removing the $10,000$ articles $d$ simultaneously. The new news corpus $D - d + E$ serves as our retrieval corpus, and we expect the model to use the events and relevant information from the retrieval corpus to generate a summary of the articles $d$.


\subsection{Text continuation: RAG Application in "Create"}

RAG is useful not only for "Delete", where it retrieves and summarizes key information from massive texts, but also for "Create". In this scenario, RAG systems show strong creativity by expanding existing texts, and we take the text continuation task as an evaluation. The text continuation task aims to automatically produce coherent and relevant subsequent content based on the beginning of the text, making the text more complete and vivid.

To construct the continuation task dataset, we follow the same method as the summary task dataset. Figure \ref{fig:add_delete} shows the construction process of text continuation. Specifically, we select a news article from a high-quality corpus and use a specialized Chinese word segmentation tool, to split it into sentences. Then, we divide the article into two equal parts: the first half serves as the input, and the second half as the output of the continuation dataset. We expect the model to use RAG technology to retrieve relevant information from the document library and generate a continuation that is coherent, informative, and consistent with the input and output.

To ensure that the retrieval database covers the real continuation text, we use the Baidu search engine to find external documents and add them to the database. The continuation text differs from the event text in that it consists of multiple sentences. Therefore, we split the continuation text into paragraphs by sentences and retrieve relevant documents for each paragraph using the search engine. This way, we guarantee that the retrieval database contains most of the information to reconstruct the continuation text.

\begin{figure}[t]
\centering
\includegraphics[width=\textwidth]{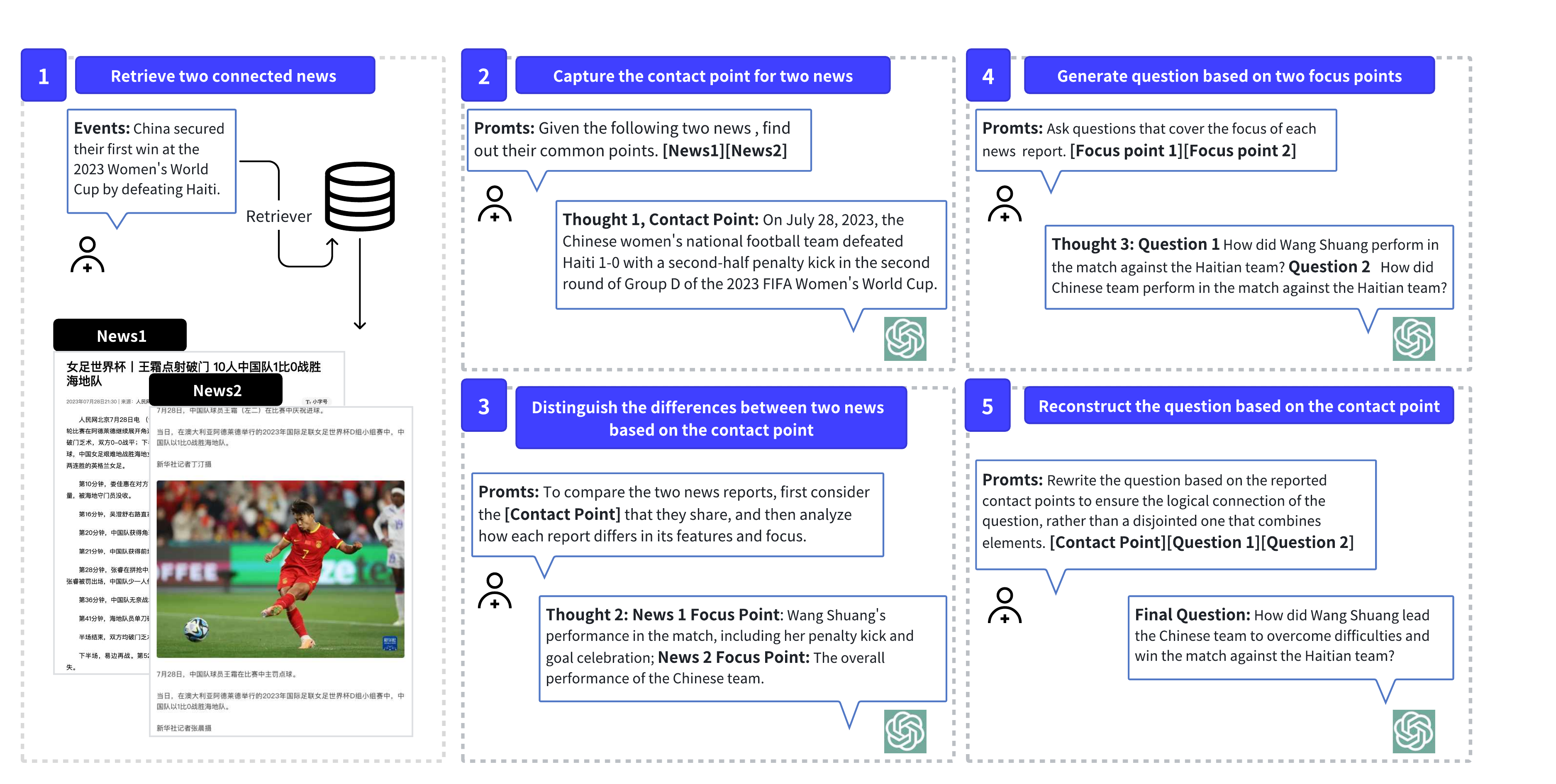}
\caption{
The dataset construction pipeline for multi-document(inferential) question answering task.
}
\label{fig:query}
\end{figure}

\subsection{Question Answering: RAG Application in "Read"}

Another application scenario of RAG is to use external knowledge bases to enhance the question-answering capabilities of LLMs, which can be applied to various knowledge-intensive tasks. Currently, there are many evaluation benchmarks to measure the performance of RAG in this scenario, and multiple question answering datasets have been created.

However, the existing question answering datasets also have some limitations. On the one hand, some datasets (such as NQ and WEBQA) are outdated, and may have been covered by LLMs in the pre-training stage, which reduces the advantage of RAG systems. On the other hand, some datasets (such as RGB) only contain some factual questions, which can be directly extracted from the retrieved texts, without requiring complex reasoning over multiple texts, which poses less challenge to RAG systems. The most recent LLMs capture enough knowledge to rival human performance across a wide variety of question answering benchmarks~\cite{bubeck2023sparks}.

To overcome these limitations, we build a large-scale question answering dataset, which is divided into two parts: single-document and multi-document question answering. Single-document question answering focuses on factual questions that ask for specific details in the news, such as the location or the main characters of an event. Multi-document question answering, on the other hand, involves inferential and critical thinking questions that require readers to reason across multiple news paragraphs, such as comparing and contrasting two events or assessing their impact.

For the single-document question answering task, we follow the dataset construction process of the previous RGB benchmark~\cite{chen2023benchmarking}. We first select news articles from our collected high-quality corpus. Then we use prompts to make GPT-4 generate questions and answers for each article. For example, for a report on "The 2023 Nobel Prize", GPT-4 will generate the question "Who was awarded the 2023 Nobel Prize for Physiology and Medicine?" and provide key information for answering it.

For the multi-document question answering task, constructing a reasoning question that requires the synthesis of multiple documents is not trivial. Simply using a prompt to force GPT-4 to generate the question is ineffective, because creating such a multi-document QA dataset is a complex reasoning task in itself.  Therefore, we adopt Chain-of-Thought (CoT) technology~\cite{wei2022chain} to enhance GPT-4. We guide the model to build the dataset gradually through multiple reasoning steps. Figure~\ref{fig:query} illustrates our specific process for building a two-document question answering dataset using GPT-4 and CoT technology. We will explain it in detail:



\begin{enumerate} 
\item \textbf{Retrieve multiple connected news}, which should cover the same event, but offer different perspectives or information. 
\item \textbf{Use prompts to help GPT-4 identify the common elements between different reports}, such as the event they report on, and ensure they are relevant.
\item \textbf{Use prompts to help GPT-4 distinguish the differences between news articles}. While keeping the connection between reports, we analyze the differences between each report. This step requires comprehensive understanding and analysis from multiple angles, and avoids generating questions that can be answered from a single paragraph.
\item \textbf{Generate the question based on different focus points}, which should require integrating information from multiple sources to answer.
\item \textbf{Reconstruct the question based on the contact point}. Based on the connections in the reports, refine the questions, ensuring the inherent logical connection, and avoiding superficial combinations. The questions should be logically linked, rather than physically juxtaposed. For example, instead of simply asking 'Describe the history of World War II and explain the basic principles of quantum physics, a question like 'How did the technological and political environment during World War II foster the development of quantum physics?' should be formulated, where the parts are interdependent or have causal relationships.
\end{enumerate}

We constructed two types of multi-document question answering datasets with different levels of difficulty: one requires reasoning from 2 documents to answer the question, and the other is more challenging and requires reasoning from 3 documents to answer the question.

To further ensure our dataset's quality, we employed a manual refinement process for the data generated by GPT-4. Our annotation team comprises three native Chinese speakers, each with at least a bachelor's degree. The annotation process is as follows: 

\begin{enumerate} 
\item The annotator evaluates the quality of the automatically generated query and chooses one of the following two options: 
\begin{itemize} 
\item \textbf{Reasonable}: Conforms to natural language usage. 
\item \textbf{Needs refinement}: Has issues with naturalness, accuracy, or grammar.
\end{itemize} 
\item If "Reasonable" is selected, no further action is taken. If "Needs refinement" is chosen, the annotator manually improves the query's naturalness and accuracy.
\end{enumerate}

In addition to their standard salary, annotators receive an extra 1 RMB per query evaluated or refined. The average annotation time per query is approximately 20 seconds. To ensure annotation quality, we randomly inspected 5\% of the annotated data.

Given the substantial cost of manual annotation and the large size of our dataset, we initially polished one-fifth of our dataset manually. We will continuously monitor dataset quality across various social media platforms and refine it manually as needed.

Notably, only 5.8\% of queries required refinement, indicating that the queries generated by GPT-4 are generally of high quality. This validates the effectiveness of using GPT-4 for initial data generation and underscores our commitment to ensuring dataset quality."

\subsection{Hallucination Modification: RAG Application in "Update"}

Besides the three scenarios mentioned above, the RAG framework can also be used to correct errors in the text. This involves using the RAG framework to access relevant information from external sources, identify and correct errors in the text, and maintain the accuracy of the text content.


We construct a hallucination modification dataset using the open-source large-scale dataset UHGEval~\cite{liang2023uhgeval}. UHGEval instructs the model to generate continuations that contain hallucinations for a given news text. It utilizes GPT-4 for automatic annotation and human evaluation to identify and mark segments in the text containing hallucinations. In our approach, we input the hallucination text along with the corresponding annotations from the dataset. Subsequently, GPT-4 is employed to rectify the hallucinations, resulting in the production of the text without any hallucinatory elements. Finally, real news continuations will be included in the document retrieval database. 

The RAG system's experimental results on this dataset can confirm if the system can retrieve the real news information from the document database based on the input text, which consists of the beginning text and the hallucination continuation text, and then correct the hallucination text to generate the text without hallucination.


\subsection{Evaluation Method}



The aim of this benchmark is to evaluate how well RAG systems can retrieve relevant documents, and use them to generate sensible responses. Therefore, we adopt an end-to-end evaluation method, which directly compares the similarity between the model output and the reference answers.

Evaluating the performance of RAG systems requires choosing appropriate evaluation metrics. We considered the previous evaluation metrics for text generation, \textbf{ROUGE} and \textbf{BLEU}, which are both based on word overlap. ROUGE mainly counts the recall rate on the n-gram, while BLEU mainly counts the precision rate on the n-gram.
However, BLEU and ROUGE are word overlap-based accuracy metrics that depend on the overall expression of the text, and do not capture the accuracy of the particular key information in the text. Therefore, they may not reflect the factual consistency of a text well, especially for long texts. To alleviate this issue, recent work~\cite{wang2020asking,scialom2021questeval,durmus2020feqa} has proposed new evaluation metrics for abstractive summarization evaluation. These metrics are based on the intuition that if you ask questions about the summary and the original document, you will get a similar answer if the summary realistically matches the original document. They evaluate the accuracy of each local piece of key information in the summary.


\begin{figure}[t]
\centering
\includegraphics[width=\textwidth]{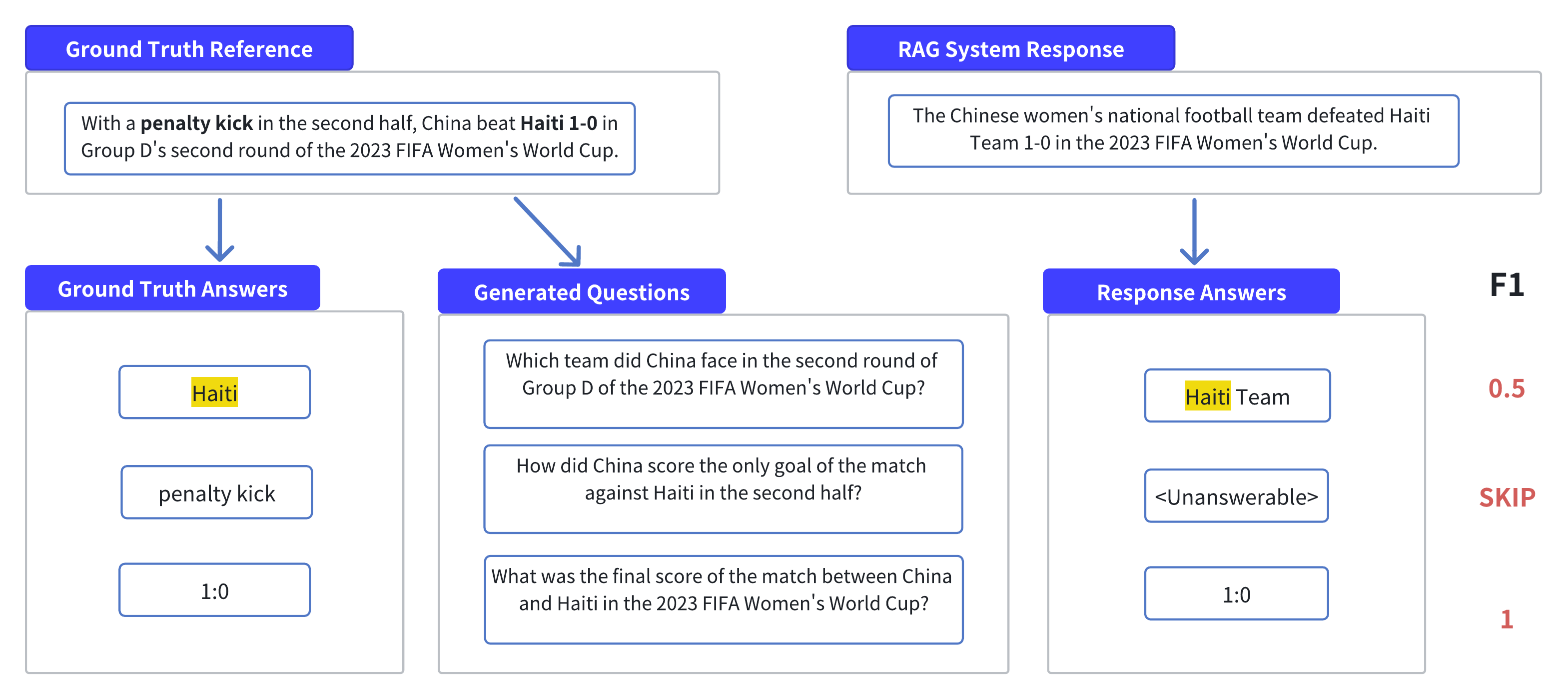}
\caption{
Overview of RAGQuestEval. A set of questions is generated based on the ground truth references. The questions are then answered using both the ground truth and the response. For the recall score of RAGQuestEval, we calculate the ratio of answerable questions to all questions(in this case, recall = 2/3). For the precision score of RAGQuestEval, corresponding answers are compared using a similarity function and averaged across questions(in this case, precision = (0.5 + 1) / 2 = 0.75).  The recall metric of RAGQuestEval indicates how much of the key information in the ground truth reference is included in the generated text, while the precision metric of RAGQuestEval indicates how correct the recalled key information is.
}
\label{fig:quest_eval}
\end{figure}

We also consider question-answering-based metrics to evaluate the factual accuracy of generation. In this paper, we examine \textbf{QuestEval}~\cite{scialom2021questeval}, a metric that improves the correlation with human judgments over previous metrics in their extensive experiments. QuestEval evaluates the factual consistency between the generated text and the source document, which is mainly used for text summarization tasks. Therefore, it does not require any ground truth reference. However, for RAG systems, the retrieved texts may be irrelevant or incorrect, so consistency with them is not a valid criterion. Instead, we use this metric to measure how well the generated text matches the ground-truth reference. We call this metric \textbf{RAGQuestEval}. We will explain this metric in detail.

Let $GT$ and $GM$ be two sequences of tokens, where  $GT$  denotes the ground truth references and $GM$ the corresponding evaluated generations. First,
we generate a series of questions from the ground truth references $GT$ using the QuestEval method, which extracts entities and noun phrases from the text. The goal of \textbf{RAGQuestEval} is to check if the generated text includes and conveys correctly all the key information from the ground truth reference.

Next, we answer these questions using both real references and model-generated text. If the question is unanswerable, the model returns "<Unanswerable>".

Finally, we calculate two scores to evaluate the quality of the generated text: recall and precision.

\paragraph{\textbf{Recall}} Recall is the ratio of answerable questions to all questions. This score shows how much information in the ground truth reference is captured by the text generated by the RAG system. A higher recall means that the generated text covers more information from the reference. 

\begin{equation}
\operatorname{Recall}(GT, GM)=\frac{1}{\left|Q_G(GT)\right|} \sum_{(q, r) \in Q_G(GT)} \mathbb{I}[Q_A(GM, q)  \neq \operatorname{<Unanswerable>}]
\end{equation}

In the above equation, $Q_G$ is the question generator and $Q_A $ is the question answerer.

\paragraph{\textbf{Precision}} Precision is the average answer similarity of all questions, excluding the unanswerable ones. We use the token level F1 score to measure the answer similarity, which is a standard metric for evaluating factoid question answering models.
Higher precision means that the generated text is more accurate and consistent with the reference. 
\begin{equation}
\operatorname{Prec}(GT, GM)=\frac{1}{\left|Q_G(GT)\right|} \sum_{(q, r) \in Q_G(GT)} F 1\left(Q_A(GM, q), r\right)
\end{equation}

\section{Experiment}
The current evaluation of RAG Benchmark only focuses on the large language model component in the RAG pipeline, and overlooks the importance of retrieval database construction and retriever. To address this gap, we examine how different aspects of RAG systems affect their performance in our benchmark. We also discuss some possible ways to improve existing RAG systems.

\subsection{Experimental Settings}
In this section, we will introduce the components of the RAG system, and describe how we conduct experiments to evaluate their impact on system performance. The RAG system consists of the following components:

\begin{itemize}
    \item \textbf{Chunk size}: The RAG system splits the external knowledge into chunks of a certain length and stores them in a vector database. The chunk size affects the retrieval accuracy and the completeness of the context.
    \item \textbf{Chunk overlap}: Chunk overlap refers to the shared tokens between two consecutive text chunks and is used to ensure semantic coherence when chunking.
    \item \textbf{Embedding model}: The RAG system converts the text chunks and the user's query into vectors using an embedding model or other methods. The embedding model affects the quality and relevance of the context.
    \item \textbf{Retriever}: The RAG system uses a retriever to find the top-k vectors most similar to the query vector in the vector database and retrieves the corresponding text chunks. The retriever affects the richness and diversity of the context.
    \item \textbf{Top-k}: This is the number of text chunks that the RAG system retrieves for each query, which serves as the context part of the LLMs prompts. The top-k influences the size of the context that the model receives.
    \item \textbf{Large language model}: The RAG system inputs the context and the query to an LLM to generate the answer. The LLM affects the correctness and rationality of the answer.
\end{itemize}

We use the following settings as the basic version of our RAG system: chunk size: 128, chunk overlap: 0\%, embedding model: bge-base, retriever: dense retriever, top-k: 8, and LLM: GPT-3.5. In the experiments, 
we change one component at a time and evaluate the results on different tasks. We compare the following values for each component:

\begin{itemize}
    \item Chunk size: 64, 128, 256, 512.
    \item Chunk overlap: 0\%, 10\%, 30\%, 50\%, 70\%.
    \item Embedding model: m3e-base, bge-base, stella-base, gte-base.
    \item Retriever: dense, bm25, hybrid, hybrid+rerank.
    \item Top-k: 2, 4, 6, 8, 10.
    \item Base LLMs: GPT-3.5, GPT-4, ChatGLM2-6B, Baichuan2-13B, Qwen-7B, Qwen-14B.
\end{itemize}

In the experiments, we use two types of evaluation metrics: The overall semantic similarity metrics(bleu, rouge-L, and bertScore) measure how closely the generated content matches the reference content in terms of meaning and fluency; and the key information metric (RAGQuestEval) measure how well the generated content captures and presents the key information from the reference content.

Considering that we used gpt-3.5 as the baseline model for the experiments, to reduce the cost, we only conducted experiments on $1/5$ of our dataset.




\begin{table}[t]
\centering
\caption{The experimental results for evaluating different chunk sizes in our benchmark. We use two types of evaluation metrics: The overall semantic similarity metrics(bleu, rouge-L, and bertScore) and the key information metric(RAGQuestEval).}
\resizebox{\textwidth}{!}{
\begin{tabular}{c|cc|cccccc}
\toprule[1pt]
\multirow{2}{*}{\textbf{task name}} & \multirow{2}{*}{\textbf{chunk size}} & \multirow{2}{*}{\textbf{topk}} & \multirow{2}{*}{\textbf{bleu}} & \multirow{2}{*}{\textbf{rouge-L}} & \multirow{2}{*}{\textbf{bertScore}} & \multicolumn{2}{c}{\textbf{RAGQuestEval}} & \multirow{2}{*}{\textbf{length}} \\
\cmidrule(lr){7-8} 
& & & & & & \textbf{precision} & \textbf{recall} & \\
\midrule

\multirow{4}{*}{text continuation} & 64 & 16 & $3.42$ & $17.67$ & $83.94$ & $26.09$ & $23.39$ & 345.8 \\
& 128 & 8 & $3.66$ & $17.78$ & $83.99$ & $26.96$ & $24.68$ & 367.6 \\
& 256 & 4 & $4.21$ & $17.93$ &\textbf{84.17}& $28.86$ & $25.99$ & 403.0 \\
& 512 & 2 & \textbf{5.12} & \textbf{18.81} & $83.57$ &\textbf{30.91}& \textbf{28.27} & 413.2 \\
\midrule

\multirow{4}{*}{summarization} & 64 & 16 & $\textbf{24.60}$ & $33.78$ & $88.07$ & $\textbf{68.29}$ & $43.98$ & 184.2  \\
& 128 & 8 & $23.69$ & $33.53$ & $88.49$ & $68.06$ & $46.18$ & 205.9 \\
& 256 & 4 & $22.97$ & $\textbf{33.85}$ & $88.83$ & $67.87$ & $48.66$ & 219.9 \\
& 512 & 2 & $21.08$ & $33.23$ & $\textbf{88.89}$ & $66.43$ & $\textbf{50.31}$ & 243.6  \\

\midrule
\multirow{4}{*}{question answering} & 64 & 16 & $37.50$ & $55.45$ & $83.02$ & $48.31$ & $68.62$ & 71.5 \\
\multirow{4}{*}{1-document}& 128 & 8 & $\textbf{39.76}$ & $\textbf{57.24}$ & $83.81$ & $52.67$ & $70.82$ & 73.3 \\
& 256 & 4 & $38.43$ & $56.20$ & $\textbf{84.02}$ & $\textbf{52.83}$ & $\textbf{72.21}$ & 79.6 \\
& 512 & 2 & $36.51$ & $54.64$ & $82.72$ & $51.26$ & $68.65$ & 84.1 \\
\midrule
\multirow{4}{*}{question answering}& 64 & 16 & $19.86$ & $34.80$ & $86.14$ & $37.77$ & $52.60$ & 143.1 \\
\multirow{4}{*}{2-document}& 128 & 8 & $22.75$ & $37.25$ & $87.16$ & $42.93$ & $56.73$ & 149.8 \\
& 256 & 4 & $\textbf{24.38}$ & $39.36$ & $88.18$ & $48.45$ & $61.75$ & 164.5 \\
& 512 & 2 & $24.05$ & $\textbf{39.69}$ & $\textbf{88.22}$ & $\textbf{49.24}$ & $\textbf{63.37}$ & 176.7   \\
\midrule
\multirow{4}{*}{question answering} & 64 & 16 & $18.55$ & $33.39$ & $86.85$ & $34.91$ & $47.95$ & 146.1  \\
\multirow{4}{*}{3-document}& 128 & 8 & $21.05$ & $35.04$ & $87.81$ & $40.32$ & $51.37$ & 156.6  \\
& 256 & 4 & $\textbf{21.63}$ & $36.03$ & $88.10$ & $42.55$ & $53.80$ &171.2  \\
& 512 & 2 & $21.40$ & $\textbf{36.55}$ & $\textbf{88.38}$ & $\textbf{44.28}$ & $\textbf{57.38}$ & 183.6  \\
\midrule
\multirow{4}{*}{hallucination}  & 64 & 16 & $\textbf{34.20}$ & $\textbf{54.90}$ & $\textbf{81.14}$ & $64.98$ & $80.96$ & 60.7 \\
\multirow{4}{*}{modification} & 128 & 8 & $32.35$ & $53.04$ & $80.49$ & $\textbf{65.07}$ & $80.85$ & 64.8 \\
& 256 & 4 & $31.48$ & $51.76$ & $80.15$ & $64.93$ & $\textbf{80.99}$ & 67.7 \\
& 512 & 2 & $30.35$ & $50.50$ & $79.66$ & $64.83$ & $79.17$ & 66.6 \\
\bottomrule[1pt]
\end{tabular}
}
\label{tab:chunk}
\end{table}

\subsection{Analyzing the Impact of Chunk Size on RAG Performance in Different Tasks}
Chunking is the process of dividing a document into chunks of a fixed length, and then converting each chunk into a vector and storing it in an index. This creates an external knowledge index. Chunk size is a crucial parameter that varies depending on the corpus characteristics. If the chunk is too small or too large, it can reduce the search accuracy or omit important content. Hence, finding the optimal chunk size is vital for ensuring search accuracy and relevance, and enabling the LLMs to generate appropriate responses. Our experiments reveal that different RAG tasks correspond to different optimal chunk sizes.

\textbf{Text Continuation:} The experimental results in Table \ref{tab:chunk} demonstrate that larger chunk size can improve the overall semantic similarity measures (bleu, rouge-L). Besides, the RAGQuestEval metrics, which reflect the precision and recall rate of key information, follow a consistent pattern. This indicates that larger blocks preserve the original document's structure, which is crucial for creative tasks such as text continuation. Smaller chunks, on the other hand, result in fragmented and semantically incoherent content, which impairs the ability of large models to understand and generate engaging content.

\textbf{Open-Domain Multi-Document Summarization: } We observe some intriguing patterns in the experimental results. Firstly, we discover that larger chunk size not only substantially increases the length of the generated text, but also cause a notable drop in the bleu score, while the rouge-L and bertScore remain almost unchanged. This implies that larger chunks can preserve more original text information, but also introduce some semantic redundancy. Secondly, for the RAGQuestEval metric that evaluates key information, we found that a larger chunk size considerably enhances the recall of key information, but also lowers the precision of key information.

We hypothesize that this is because larger chunks enable the retrieval of more relevant content, 
thus improving the recall of key information. However, larger chunks also make the summarization task more challenging, as more fine-grained selection is required from the more relevant information, leading to lower precision of key information, which may not be a good thing for the summary.

\textbf{Question Answering: } 
For single-document QA, too large chunks will reduce both recall and precision score of key information. The task only requires extracting information from a sub-paragraph of a single document, and the answer may be in a specific sentence. Therefore, smaller chunks are more suitable, as excessive content will make the extraction harder for the model.

For multi-document QA, the results are different from those of single-document QA. Larger chunks can significantly improve the recall and precision of key information, as well as the semantic similarity of the generated and reference answers. This is because larger chunks retain the original structure of the article, which is crucial for reasoning and understanding tasks, and fragmented information is not conducive to reasoning.

\textbf{Hallucination Modification}: For the hallucination modification task, the results are similar to those of the single-document QA task. Smaller chunks can significantly improve the semantic similarity metrics, such as the bleu score. This indicates that in the hallucination dataset created by UHGEval, the hallucination information often pertains to only one sentence, which is a mistake at the word or entity level, and does not require the comprehension of long text. Hence, there is no need to understand the whole document, only the relevant portions can be retrieved and modified.

\begin{table}[t]
\centering
\caption{The experimental results for evaluating different chunk overlap values in our benchmark.}
\resizebox{\textwidth}{!}{
\begin{tabular}{c|c|cccccc}
\toprule[1pt]
\multirow{2}{*}{\textbf{task name}} &\multirow{2}{*}{\textbf{chunk overlap(\%)}}& \multirow{2}{*}{\textbf{bleu}} & \multirow{2}{*}{\textbf{rouge-L}} & \multirow{2}{*}{\textbf{bertScore}} & \multicolumn{2}{c}{\textbf{RAGQuestEval}} & \multirow{2}{*}{\textbf{length}} \\
\cmidrule(lr){6-7}
& & & & & \textbf{precision} & \textbf{recall} & \\ 

\midrule 
\multirow{5}{*}{text continuation}  & 0 & $3.66 $ & $17.78 $ & $83.99 $ & $26.96 $ & $24.68 $ & 367.6 \\
& 10 & $3.86 $ & $17.84 $ & $84.03 $ & $27.18 $ & $24.21 $ & 359.2 \\ 
& 30 & $3.91 $ & $17.92 $ & $\textbf{84.12} $ & $28.21 $ & $24.72 $ & 367.0 \\
& 50 & $3.94 $ & $17.86 $ & $84.01 $ & $\textbf{28.34 }$ & $24.48 $ & 365.4 \\ 
& 70 & $\textbf{4.03 }$ & $\textbf{17.95} $ & $84.04 $ & $27.64 $ & $\textbf{25.32} $ & 364.0 \\
\midrule 

\multirow{5}{*}{summarization} & 0 & $23.69 $ & $33.53 $ & $88.49 $ & $68.06 $ & $46.18 $ & 205.9 \\
& 10 & $23.54 $ & $33.59 $ & $88.35 $ & $\textbf{68.67} $ & $46.16 $ & 208. 4 \\
& 30 & $23.74 $ & $33.58 $ & $88.41 $ & $68.02 $ & $46.08 $ & 203.3 \\
& 50 & $24.05 $ & $33.99 $ & $88.62 $ & $68.61 $ & $46.64 $ & 204.2 \\
& 70 & $\textbf{24.49} $ & $\textbf{34.29} $ & $\textbf{88.71} $ & $68.45 $ & $\textbf{47.08} $ & 201.8 \\

\midrule 

\multirow{5}{*}{question answering} & 0 & $\textbf{39.76} $ & $57.24 $ & $83.81 $ & $52.67 $ & $70.82 $ & 73. 3 \\
\multirow{5}{*}{1-document} & 10 & $39.36 $ & $\textbf{57.59} $ & $83.77 $ & $51.87 $ & $71.36 $ & 73. 3 \\
& 30 & $39.43 $ & $57.40 $ & $83.87 $ & $53.30 $ & $72.74 $ & 73.5 \\
& 50 & $39.31 $ & $57.27 $ & $\textbf{84.14 }$ & $53.85 $ & $73.63 $ & 74.6 \\
& 70 & $38.46 $ & $57.01 $ & $84.10 $ & $\textbf{54.06} $ & $\textbf{73.94} $ & 75.5 \\
\midrule 
\multirow{5}{*}{question answering} & 0 & $22.75 $ & $37.25 $ & $87.16 $ & $42.93 $ & $56.73 $ & 149.8 \\
\multirow{5}{*}{2-document} & 10 & $23.41 $ & $37.72 $ & $87.33 $ & $43.18 $ & $56.50 $ & 149.4 \\
 & 30 & $23.02 $ & $37.37 $ & $87.24 $ & $43.64 $ & $58.25 $ & 149.4 \\
& 50 & $23.65 $ & $38.33 $ & $87.61 $ & $43.98 $ & $59.21 $ & 152.2 \\
& 70 & $\textbf{23.69} $ & $\textbf{38.51} $ & $\textbf{87.76} $ & $\textbf{44.84 }$ & $\textbf{59.53} $ & 152.2 \\
\midrule 

\multirow{5}{*}{question answering} & 0 & $21.05 $ & $35.04 $ & $87.81 $ & $40.32 $ & $51.37 $ & 156.6 \\
\multirow{5}{*}{3-document} & 10 & $21.08 $ & $\textbf{35.56} $ & $87.57 $ & $41.62 $ & $50.74 $ & 154.6 \\
& 30 & $21.39 $ & $35.49 $ & $87.78 $ & $40.96 $ & $51.33 $ & 155.9 \\
& 50 & $\textbf{21.60} $ & $35.48 $ & $87.83 $ & $\textbf{41.91} $ & $\textbf{51.97} $ & 157.4 \\
& 70 & $21.10 $ & $35.11 $ & $\textbf{87.95 }$ & $41.39 $ & $51.58 $ & 158.9 \\

\midrule 

\multirow{5}{*}{hallucination} & 0 & $32.35 $ & $53.04 $ & $80.49 $ & $65.07 $ & $80.85 $ & 64.8 \\
\multirow{5}{*}{modification} & 10 & $32.57 $ & $53.29 $ & $80.51 $ & $65.30 $ & $\textbf{81.36} $ & 63.9 \\
& 30 & $\textbf{33.72 }$ & $\textbf{53.98} $ & $\textbf{80.69} $ & $64.53 $ & $80.91 $ & 63.6 \\
& 50 & $32.58 $ & $52.92 $ & $80.49 $ & $65.07 $ & $80.18 $ & 65.7 \\
& 70 & $31.77 $ & $52.13 $ & $80.12 $ & $\textbf{65.80} $ & $81.06 $ & 66.9 \\

\bottomrule[1pt]
\end{tabular}
 }
\label{tab:chunk_overlap}
\end{table}

\subsection{Analyzing the Impact of Chunk Overlap on RAG Performance in Different Tasks}
Chunk overlap is the number of tokens that two adjacent chunks share. To keep the text semantics coherent, adjacent chunks have some overlap. Chunk overlap determines the size of this overlap. This splitting method meets the maximum length limit of LLMs and maintains the semantic connection between adjacent chunks. Suitable chunk size and overlap can enhance the fluency and coherence of LLMs for long texts. We will show how the chunk overlap rate affects the system performance for different tasks in Table \ref{tab:chunk_overlap}.

\textbf{Text Continuation:} With an increase in chunk overlap, we observe a slight enhancement in the metrics that evaluate the alignment of generated text with a reference answer (bleu, rouge-L, and bertScore). The RAGQuestEval metric, which evaluates the accuracy and completeness of important information, improves more obviously. These results indicate that a greater chunk overlap is beneficial for preserving the flow of ideas in the text, which is essential for tasks that require generating new, creative content.

\textbf{Open-Domain Multi-Document Summarization:} During summarization tasks, all evaluation metrics show a slight improvement as chunk overlap grows. Interestingly, despite assumptions that more overlap might reduce the variety of context information available, this does not result in a lower rate of recalling important information. In fact, the best performance in terms of recall occurs at a chunk overlap of 70\%. This could mean that a larger overlap allows the model to focus more on the main points and ignore less relevant or redundant information.

\textbf{Question Answering:}  In question answering tasks, chunk overlap has minimal impact on overall semantic similarity metrics such as bleu, rouge-L, and bertScore. However, it significantly affects the accuracy and recall metrics for key information. The results indicate that as chunk overlap increases, the accuracy and recall of key information in single-document question answering tasks improve substantially. Similar improvements are observed in two-document question answering tasks. However, for three-document question answering tasks, the improvement is less pronounced. This may be because three-document question answering tasks require richer context, and larger chunk overlaps may reduce the available context.


\textbf{Hallucination  Modification:} Changes in chunk overlap have a minimal effect on the performance metrics for tasks that involve correcting hallucinations. This is likely due to the errors in these tasks typically being specific to individual entities or words, making the consistency of the chunks less impactful.

\begin{table}[t]
\centering
\caption{The experimental results for evaluating different retrievers in our benchmark.}
\resizebox{\textwidth}{!}{
\begin{tabular}{c|c|cccccc}
\toprule[1pt]
\multirow{2}{*}{\textbf{task name}} &\multirow{2}{*}{\textbf{retriever name}}& \multirow{2}{*}{\textbf{bleu}} & \multirow{2}{*}{\textbf{rouge-L}} & \multirow{2}{*}{\textbf{bertScore}} & \multicolumn{2}{c}{\textbf{RAGQuestEval}} & \multirow{2}{*}{\textbf{length}} \\
\cmidrule(lr){6-7}
& & & & & \textbf{precision} & \textbf{recall} & \\
\midrule 
\multirow{4}{*}{text continuation} & BM25 & $3.51$ & $17.56$ & $83.83$ & $\textbf{27.25}$ & $23.70$ & 370.5 \\
& Dense & $3.66$ & $\textbf{17.78}$ & $\textbf{83.99}$ & $26.96$ & $\textbf{24.68}$ & 367.6 \\
& Hybrid & $\textbf{3.69}$ & $17.69$ & $83.97$ & $27.24$ & $24.01$ & 362.4 \\
& Hybrid+Rerank & $3.55$ & $17.55$ & $83.90$ & $26.69$ & $24.02$ & 370.3 \\

\midrule 
\multirow{4}{*}{summarization} & BM25 & $\textbf{25.19}$ & $33.77$ & $87.82$ & $\textbf{70.78}$ & $44.30$ & 190.4 \\
& Dense & $23.69$ & $33.53$ & $\textbf{88.49}$ & $68.06$ & $46.18$ & 205.9 \\
& Hybrid & $24.21$ & $33.81$ & $88.24$ & $68.70$ & $45.63$ & 199.8  \\
& Hybrid+Rerank & $24.33$ & $\textbf{33.90}$ & $88.48$ & $68.34$ & $\textbf{46.41}$ & 200.2 \\

\midrule 
\multirow{4}{*}{question answering} & BM25 & $39.91$ & $57.33$ & $83.36$ & $51.90$ & $69.17$ & 69.6 \\
\multirow{4}{*}{1-document} & Dense & $39.76$ & $57.24$ & $83.81$ & $52.67$ & $70.82$ & 73.3 \\
& Hybrid & $39.67$ & $57.38$ & $84.06$ & $52.71$ & $70.83$ & 70.8 \\
& Hybrid+Rerank & $\textbf{40.63}$ & $\textbf{58.26}$ & $\textbf{84.68}$ & $\textbf{54.60}$ & $\textbf{73.92}$ & 72.8 \\

\midrule 
\multirow{4}{*}{question answering} & BM25 & $\textbf{24.61}$ & $38.31$ & $86.86$ & $42.26$ & $54.56$ & 138.4 \\
\multirow{4}{*}{2-document}& Dense & $22.75$ & $37.25$ & $87.16$ & $42.93$ & $56.73$ & 149.8 \\
& Hybrid & $24.03$ & $38.43$ & $87.30$ & $45.67$ & $58.01$ & 144.6 \\
& Hybrid+Rerank & $24.53$ & $\textbf{38.91}$ & $\textbf{87.89}$ & $\textbf{47.18}$ & $\textbf{58.12}$ & 151.7 \\

\midrule 
\multirow{4}{*}{question answering} & BM25 & $20.98$ & $34.33$ & $87.02$ & $37.04$ & $48.53$ & 147.6 \\
\multirow{4}{*}{3-document}& Dense & $21.05$ & $35.04$ & $87.81$ & $40.32$ & $51.37$ & 156.6 \\
& Hybrid & $21.35$ & $35.34$ & $87.66$ & $41.07$ & $51.09$ & 150.8 \\
& Hybrid+Rerank & $\textbf{21.74}$ & $\textbf{35.88}$ & $\textbf{88.21}$ & $\textbf{41.59}$ & $\textbf{52.84}$ & 157.1 \\

\midrule 
\multirow{4}{*}{hallucination} & BM25 & $\textbf{33.09}$ & $\textbf{54.21}$ & $\textbf{80.86}$ & $64.80$ & $79.90$ & 59.0 \\
\multirow{4}{*}{modification} & Dense & $32.35$ & $53.04$ & $80.49$ & $65.07$ & $80.85$ & 64.8 \\
& Hybrid & $32.22$ & $52.92$ & $80.57$ & $\textbf{66.30}$ & $\textbf{81.03}$ & 63.4 \\
& Hybrid+Rerank & $32.62$ & $53.01$ & $80.62$ & $65.57$ & $80.82$ & 64.9 \\

\bottomrule[1pt]
\end{tabular}
 }
\label{tab:retriever}
\end{table}

\begin{figure}[t]
\centering
\includegraphics[width=0.8\linewidth]{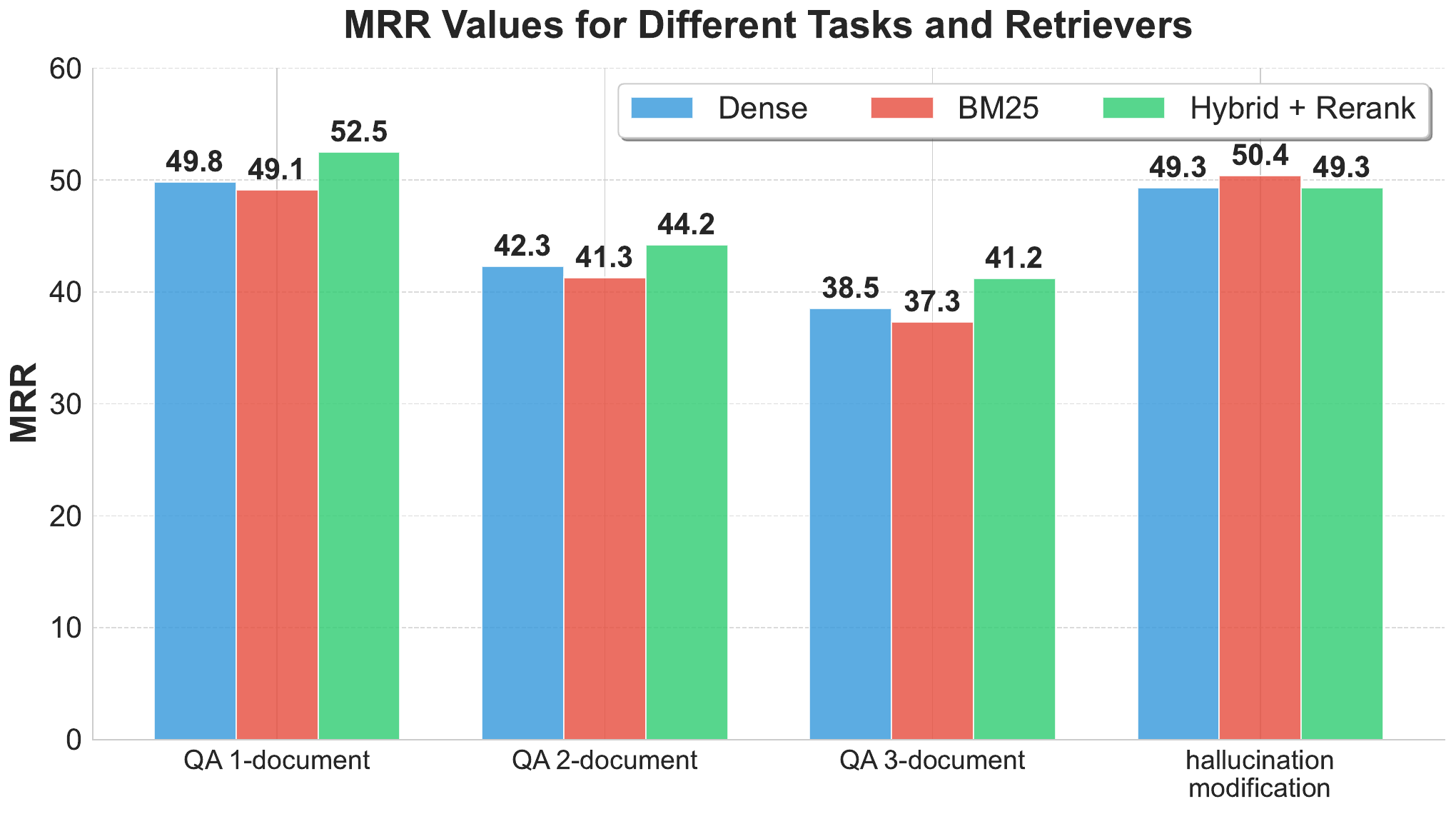}
\caption{Comparison of Mean Reciprocal Rank (MRR) scores for different retrieval methods in our benchmark.}
\label{fig:mrr_retriever}
\end{figure}

\begin{table}[t]
\centering
\caption{The experimental results for evaluating different embedding models in our benchmark.}
\resizebox{\textwidth}{!}{
\begin{tabular}{c|c|cccccc}
\toprule[1pt]
\multirow{2}{*}{\textbf{task name}} &\multirow{2}{*}{\textbf{embedding name}}& \multirow{2}{*}{\textbf{bleu}} & \multirow{2}{*}{\textbf{rouge-L}} & \multirow{2}{*}{\textbf{bertScore}} & \multicolumn{2}{c}{\textbf{RAGQuestEval}} & \multirow{2}{*}{\textbf{length}} \\
\cmidrule(lr){6-7} 
& & & & & \textbf{precision} & \textbf{recall} & \\
\midrule 
\multirow{4}{*}{text continuation} & m3e-base & $3.59 $ & $17.55 $ & $83.76 $ & $27.30 $ & $23.73 $ & 350.0 \\

& bge-base & $3.66 $ & $17.78 $ & $83.99 $ & $26.96 $ & $\textbf{24.68}$ & 367.6 \\
& stella-base & $3.73 $ & $17.67 $ & $\textbf{84.05} $ & $\textbf{28.78}$ & $24.65 $ & 366.6 \\
& gte-base & $\textbf{3.76} $ & $\textbf{17.80} $ & $84.03 $ & $27.35 $ & $24.18 $ & 362.1 \\

\midrule 

\multirow{4}{*}{summarization} & m3e-base & $22.91 $ & $33.23 $ & $88.31 $ & $68.58 $ & $46.02 $ & 210.5 \\
& bge-base & $23.69 $ & $\textbf{33.53} $ & $88.49 $ & $68.06 $ & $46.18 $ & 205.9 \\
& stella-base & $\textbf{23.50} $ & $ 33.50 $ & $88.58 $ & $\textbf{68.22} $ & $46.56 $ & 205.5 \\
& gte-base & $22.87 $ & $33.46 $ & $\textbf{88.58} $ & $68.10 $ & $\textbf{47.13} $ & 211.1 \\

\midrule 
\multirow{4}{*}{question answering}& m3e-base & $38.81 $ & $56.49 $ & $83.41 $ & $50.18 $ & $69.72 $ & 75.2 \\

\multirow{4}{*}{1-document} & bge-base & $\textbf{39.76} $ & $57.24 $ & $83.81 $ & $52.67 $ & $70.82 $ & 73.3 \\
& stella-base & $39.58 $ & $\textbf{57.28} $ & $\textbf{83.91} $ & $\textbf{53.13} $ & $71.74 $ & 73.9 \\
& gte-base & $39.58 $ & $57.19 $ & $83.90 $ & $52.39 $ & $\textbf{71.97} $ & 76.5 \\
\midrule 
\multirow{4}{*}{question answering} & m3e-base & $22.32 $ & $36.81 $ & $86.91 $ & $42.97 $ & $55.67 $ & 148.4 \\
\multirow{4}{*}{2-document} & bge-base & $22.75 $ & $37.25 $ & $87.16 $ & $42.93 $ & $56.73 $ & 149.8 \\
& stella-base & $\textbf{23.39} $ & $\textbf{37.75} $ & $ 87.37 $ & $\textbf{44.83} $ & $\textbf{58.00} $ & 149.5 \\
& gte-base & $23.20 $ & $37.59 $ & $\textbf{87.48} $ & $43.99 $ & $57.58 $ & 151.5 \\
\midrule 
\multirow{4}{*}{question answering}& m3e-base & $20.72 $ & $34.78 $ & $87.43 $ & $39.57 $ & $50.88 $ & 154.3 \\
\multirow{4}{*}{3-document}& bge-base & $21.05 $ & $35.04 $ & $87.81 $ & $40.32 $ & $\textbf{51.37}$ & 156.6 \\
& stella-base & $\textbf{21.26} $ & $35.27 $ & $\textbf{87.81} $ & $\textbf{41.4}1 $ & $50.42 $ & 154.4 \\
& gte-base & $21.15 $ & $\textbf{35.59} $ & $87.86 $ & $40.18 $ & $51.11 $ & 157.2 \\

\midrule 

\multirow{4}{*}{hallucination} & m3e-base & $\textbf{32.83} $ & $\textbf{53.27} $ & $\textbf{80.78} $ & $\textbf{65.87} $ & $\textbf{81.69} $ & 64.5 \\

\multirow{4}{*}{modification}& bge-base & $32.35 $ & $53.04 $ & $80.49 $ & $65.07 $ & $80.85 $ & 64.8 \\
& stella-base & $32.34 $ & $52.96 $ & $80.59 $ & $65.74 $ & $81.50 $ & 65.2 \\

& gte-base & $31.69 $ & $52.46 $ & $80.40 $ & $65.35 $ & $80.69 $ & 64.5 \\

\bottomrule[1pt] 
\end{tabular}
}
\end{table}

\subsection{Analyzing the Impact of Retriever on RAG Performance in Different Tasks}


A retriever is a key component of the RAG pipeline, which finds relevant documents from a large database based on the user input, and provides contextual information for the large model. There are two main types of retrievers:
\textbf{Keyword-based search-sparse retrieval algorithms}, which use keywords and their frequencies to compute the relevance between documents and queries. Common sparse retrieval algorithms include TF-IDF and BM25. BM25 is an enhanced TF-IDF method, which accounts for factors such as the length and position of words in the document.
\textbf{Dense retrieval algorithms}, which use deep learning models to encode documents and queries into low-dimensional vectors, and then measure the cosine similarity between them. This method can capture the semantic and contextual information of words, and improve the retrieval performance.

In order to combine the advantages of both types of retrievers, we can fuse their retrieval results and randomly sample k from them as contexts for LLMs(\textbf{Hybrid}). Alternatively, we can also use a re-ranking model to re-rank the fused retrieval results, and then select the top-k ones as the context of LLMs(\textbf{Hybrid+Rerank}). In our experiments, we employ the bge-rank as the rerank model.

\textbf{Text Continuation: } As Table \ref{tab:retriever} displays, the performance of the dense retriever is roughly equivalent to that of BM25, except for the key information recall rate. Compared to the keyword-based algorithm, the modern vector search can capture the semantic and contextual information of words, so that more content that does not match keywords but is obviously semantically related can be retrieved.
However, the RAG system using BM25 also performs well. In terms of the precision of key information, BM25 even exceeds the dense retriever. This suggests that in the continuation task, which is a creative task, BM25 can retrieve content that is highly relevant to the user's intention, but may overlook some details.

\textbf{Open-Domain Multi-Document Summarization:} On the overall semantic similarity metric, the performance of the dense retriever is roughly equivalent to that of BM25. On the QuestEval metric, BM25 surpasses dense retriever in terms of key information precision, but slightly trails behind in key information recall. If the retrieved content contains a lot of irrelevant information, the model-generated summary may have errors or redundancies. BM25 retrieved content usually matches the user's intention better, but sometimes may miss some important information. Therefore, BM25 is weaker than dense retriever in key information recall, but excels in key information accuracy. Besides, hybrid retrieval algorithms presumably combine the advantages of both, and the RAG system generates content with suitable precision and recall.

\textbf{Question Answering: } 
In question answering, 
we find that dense retriever has a more obvious advantage over BM25, 
when dealing with reasoning questions that require synthesizing multiple documents. In question-answering tasks that require considering three documents, Dense retriever not only surpasses BM25 in all the overall semantic similarity metrics, but also achieves a significant improvement in key information precision and recall. This indicates that question-answering retrieval is more difficult than text continuation and other tasks, especially reasoning question-answering, which requires a higher level of semantic understanding, and simple keyword retrieval algorithms may not be sufficient. We also found that the Hybrid+Rerank algorithm, which combines and re-ranks the results of both algorithms, improves on all evaluation metrics. This suggests that this is a better retrieval algorithm for question-answering tasks.

\textbf{Hallucination Modification: } 
Consistent with the conclusion of summarization, the BM25 retriever performs slightly better than or equal to the dense retriever.
For RAG tasks such as hallucination modification, which require precise retrieval of highly relevant content, BM25 shows good performance. Moreover, BM25 requires less computational resources than dense retrievers. This indicates that different RAG tasks require different retrieval algorithms.

\textbf{Retrieval Accuracy Evaluation: }  To make a more comprehensive evaluation, we evaluated the retrieval accuracy on question answering and hallucination modification tasks using MRR (mean reciprocal rank) as a separate metric. This separate evaluation allows for a more accurate assessment of the retriever's capabilities. Notably, text continuation and open-domain summarization tasks were excluded due to their subjective and vague evaluation criteria, lacking clear ground truth. Additionally, both 2-document and 3-document question answering require multiple documents to address queries. Therefore, we calculate the MRR for each retrieved document individually and take the average as the final result. The pure hybrid algorithm was not evaluated separately as it could alter the order of retrieved content, affecting subsequent processing steps.

Figure \ref{fig:mrr_retriever} shows that the hybrid + reranking method excels in most tasks, outperforming other methods. This demonstrates the effectiveness of combining multiple retrieval strategies with reranking. Notably, BM25 and dense retrievers perform comparably in many cases, highlighting the strengths of both traditional and neural network methods. In question answering, performance for all methods declines as the number of documents increases, aligning with expectations since multi-document tasks are more challenging and require stronger information integration. These results are consistent with our previous end-to-end evaluations, confirming the reliability of the end-to-end evaluation method.

\begin{figure}[t]
\centering
\includegraphics[width=0.8\linewidth]{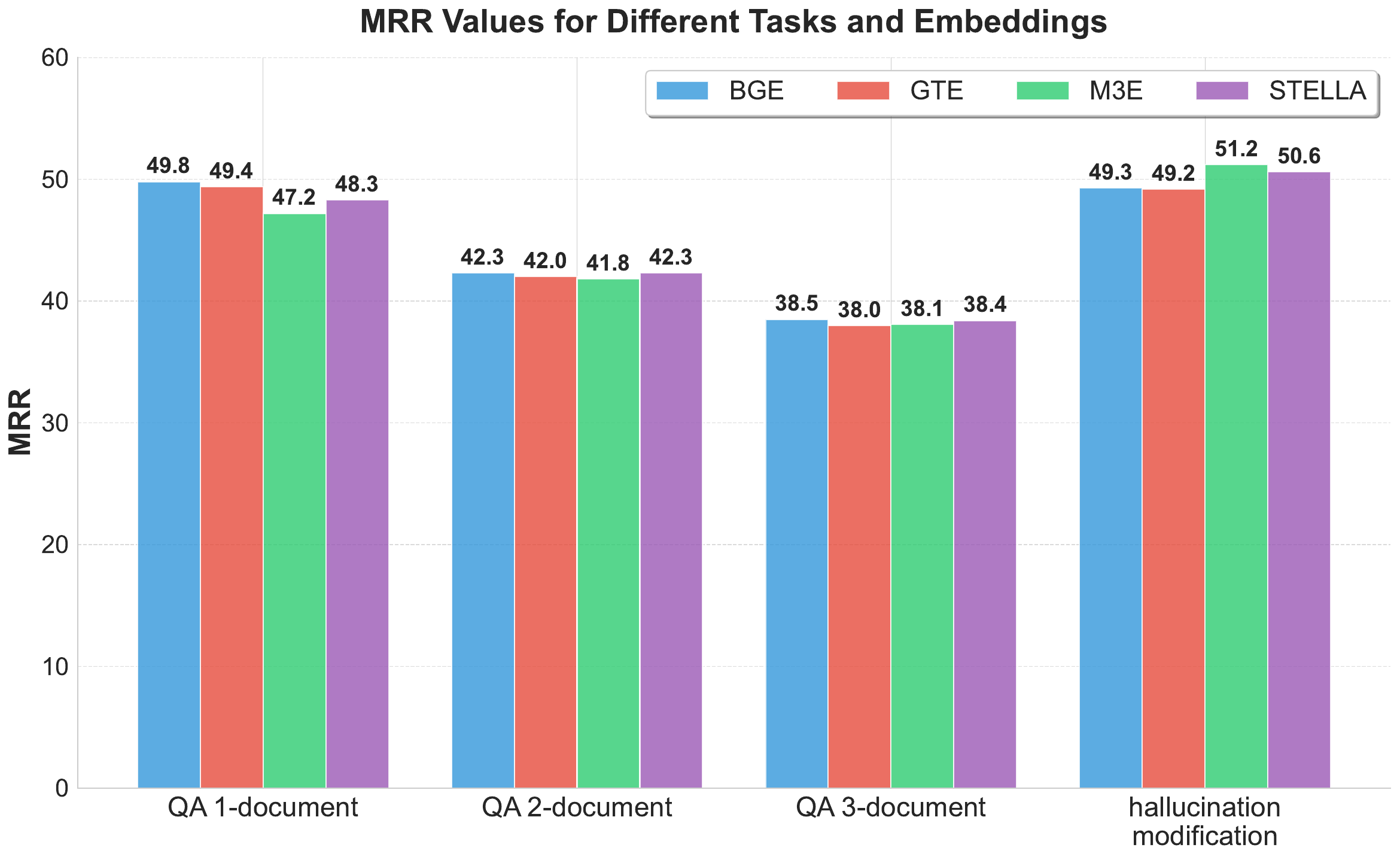}
\caption{Comparison of Mean Reciprocal Rank (MRR) scores for different embedding models in our benchmark.}
\label{fig:mrr_embedding}
\end{figure}

\subsection{Analyzing the Impact of Embedding Model on RAG Performance in Different Tasks}
Most RAG systems use vector similarity-based algorithms as retrievers. Therefore, the embedding model that converts document blocks into vectors is crucial for the retrieval effect. We tested various embedding models optimized for retrieval tasks and compared their performance in the RAG system.
We evaluated several embedding models with similar parameter sizes. According to ~\cite{muennighoff2022mteb}, the embedding models' performance on the retrieval task should follow the order of GTE > STELLA > BGE > M3E. Our results show some variations with this order.

For creative tasks like continuation, the relevance of the retrieved content was often ambiguous. Thus, we noticed that the performance difference between the embedding models was small.

For single-document question answering tasks that required precise localization of relevant documents, we found that m3e-base performed much worse than others. This matched the finding of ~\cite{muennighoff2022mteb}. However, for the hallucination modification task, m3e-base, which ranked the lowest on the retrieval benchmark, outperformed the other models on all metrics.  These results further show that the retrieval benchmark may not be fully appropriate for RAG. 



\textbf{Retrieval Accuracy Evaluation: }  Similar to the experiments in the retriever evaluation, we use the MRR metric to evaluate four mainstream embedding methods: BGE, GTE, M3E, and STELLA. The results in Figure \ref{fig:mrr_embedding} indicate that the performance of these methods is relatively close across different tasks, with no single method outperforming the others in all tasks. This underscores the importance of considering specific task requirements when selecting an embedding method.

As the number of documents increases from 1 to 3, the MRR values for all methods show a downward trend. This trend aligns with our previous end-to-end experimental results, highlighting the challenges of multi-document understanding tasks.

\begin{table}[t]
\centering
\caption{The experimental results for evaluating different top-k values in our benchmark.}
\resizebox{\textwidth}{!}{
\begin{tabular}{c|c|ccccccc}
\toprule[1pt]
\multirow{2}{*}{\textbf{task name}}  & \multirow{2}{*}{\textbf{topk}}  & \multirow{2}{*}{\textbf{bleu}}  & \multirow{2}{*}{\textbf{rouge-L}}  & \multirow{2}{*}{\textbf{bertScore}}  & \multicolumn{2}{c}{\textbf{RAGQuestEval}} & \multirow{2}{*}{\textbf{length}}  \\
\cmidrule(lr){6-7}
& & & & & \textbf{precision} & \textbf{recall} & \\
\midrule
\multirow{5}{*}{text continuation}  & 2 & $2.89$ & $17.20$ & $83.60$ & $25.35$ & $23.14$ & 367.0 \\
&  4 & $3.34$ & $17.49$ & $83.80$ & $26.66$ & $23.54$ & 369.3 \\
& 6 & $3.53$ & $17.64$ & $83.81$ & $\textbf{27.66}$ & $24.32$ & 375.4 \\
&  8 & $3.66$ & $17.78$ & $83.99$ & $26.96$ & $24.68$ & 367.6 \\
& 10 & $\textbf{3.91}$ & $\textbf{17.84}$ & $\textbf{84.01}$ & $27.61$ & $\textbf{25.00}$ & 355.7 \\
\midrule
\multirow{5}{*}{summarization} & 2 & $\textbf{26.86}$ & $\textbf{33.87}$ & $87.34$ & $\textbf{70.08}$ & $42.21$ & 161.0 \\
 & 4 & $24.78$ & $33.62$ & $87.95$ & $68.91$ & $44.19$ & 185.6 \\
 & 6 & $23.71$ & $33.36$ & $88.16$ & $68.28$ & $45.08$ & 198.6 \\
 & 8 & $23.69$ & $33.53$ & $88.49$ & $68.06$ & $46.18$ & 205.9 \\
 & 10 & $23.62$ & $33.56$ & $\textbf{88.51}$ & $68.17$ & $\textbf{46.70}$ & 208.3 \\
\midrule
\multirow{5}{*}{question answering} & 2 & $39.13$ & $56.26$ & $82.57$ & $50.81$ & $65.80$ & 67.7 \\
\multirow{5}{*}{1-document} & 4 & $39.47$ & $56.58$ & $83.39$ & $52.14$ & $69.53$ & 70.6 \\
& 6 & $39.40$ & $56.86$ & $83.81$ & $52.60$ & $70.80$ & 72.5 \\
& 8 & $\textbf{39.76}$ & $\textbf{57.24}$ & $83.81$ & $52.67$ & $\textbf{70.82}$ & 73.3 \\
& 10 & $38.84$ & $56.52$ & $\textbf{83.93}$ & $\textbf{53.67}$ & $70.31$ & 74.1 \\
\midrule
\multirow{5}{*}{question answering} & 2 & $21.65$ & $35.16$ & $84.72$ & $36.91$ & $47.41$ & 126.5 \\
\multirow{5}{*}{2-document} & 4 & $22.33$ & $36.68$ & $86.39$ & $41.15$ & $52.78$ & 139.5 \\
& 6 & $\textbf{23.04}$ & $37.43$ & $87.01$ & $43.29$ & $55.47$ & 143.7 \\
& 8 & $22.75$ & $37.25$ & $87.16$ & $42.93$ & $56.73$ & 149.8 \\
& 10 & $22.90$ & $\textbf{37.63}$ & $\textbf{87.43}$ & $\textbf{43.88}$ & $\textbf{57.34}$ & 153.4 \\
\midrule
\multirow{5}{*}{question answering}  & 2 & $19.27$ & $32.57$ & $85.65$ & $33.70$ & $43.90$ & 136.3 \\
\multirow{5}{*}{3-document}& 4 & $20.23$ & $34.21$ & $86.93$ & $37.26$ & $48.35$ & 145.5 \\
&6 & $20.73$ & $34.95$ & $87.66$ & $39.59$ & $51.03$ & 151.3 \\
& 8 & $\textbf{21.05}$ & $\textbf{35.04}$ & $87.81$ & $40.32$ & $51.37$ & 156.6 \\
& 10 & $20.61$ & $35.01$ & $\textbf{88.02}$ & $\textbf{40.90}$ & $\textbf{52.11}$ & 162.5 \\
\midrule
\multirow{5}{*}{hallucination}  & 2 & $32.12$ & $53.00$ & $\textbf{80.54}$ & $64.95$ & $79.24$ & 59.6 \\
\multirow{5}{*}{modification} & 4 & $\textbf{32.50}$ & $52.94$ & $80.53$ & $\textbf{65.18}$ & $79.34$ & 60.2 \\
& 6 & $32.32$ & $52.70$ & $80.36$ & $64.48$ & $79.27$ & 61.8 \\
& 8 & $32.35$ & $\textbf{53.04}$ & $80.49$ & $65.07$ & $80.85$ & 64.8 \\
& 10 & $31.30$ & $51.71$ & $80.09$ & $64.84$ & $\textbf{80.90}$ & 68.3 \\
\bottomrule[1pt]
\end{tabular}
}
\label{tab:topk}
\end{table}

\subsection{Analyzing the Impact of Top-k on RAG Performance in Different Tasks}
The RAG system converts the user's query into a vector using the same embedding model as the vector database. Then, it searches the index for the top-k most similar vectors to the query vector, and retrieves the corresponding text blocks from the database. These text blocks serve as the context for the LLM prompt. The amount of information that the model receives depends on the size of k. We will show how the amount of context information affects the system performance for different tasks in Table \ref{tab:topk}.


\textbf{Text Continuation}: Text continuation is a highly creative task. Table \ref{tab:topk} shows that increasing top-k improves both the overall semantic similarity metrics (bertScore, bleu, and rouge-L) and the RAGQuestEval metrics. The recall metric of RAGQuestEval shows how much key information from the reference is included in the generated text, while the precision metric shows how correct and relevant that information is. We found that higher top-k values lead to higher recall and precision scores, indicating that the generated text contains more and better key information. We attribute this to the increased diversity and accuracy of the generated text from more documents.

\textbf{Open-Domain Multi-Document Summarization: } Increasing the top-k value leads to longer and lower-quality summaries. The rouge-L and bertScore metrics stay almost the same, but the bleu metric drops significantly, indicating less similarity between the summaries and the references. The top-k value also affects the key information metrics. Higher top-k values increase the recall scores, meaning more key information is included, but decrease the precision scores, meaning more errors or redundancies are present.


\textbf{Question Answering:} For single-document QA, increasing top-k has little impact on the semantic similarity metric, but improves the RAGQuestEval metrics, which measure the accuracy and recall of key information. When the top-k value is too small, increasing the value of the top-k can significantly increase the recall and precision scores. This is because when the retrieved content is small, it may not be helpful for answering.

For multi-document QA (2-document and 3-document), increasing top-k significantly improves the recall and precision scores, as there are more chances to retrieve two relevant and complementary documents. More documents can also provide additional information, which helps to bridge the knowledge gap between documents and give more comprehensive answers. The results of 2-document and 3-document question answering are similar.


\textbf{Hallucination Modification: } The top-k value has little effect on the semantic similarity metrics (bleu, rouge and bertScore) and the key information metric (RAGQuestEval). They only drop sharply when the top-k is too large. This is because, in our hallucination modification dataset, correcting the wrong information only requires a small amount of context, and the model has a certain anti-interference ability in the hallucination modification task, so the top-k value is not a decisive factor.

\begin{table}[t]
\centering
\caption{The experimental results for evaluating different large language models in our benchmark.}
\resizebox{0.86\textwidth}{!}{
\begin{tabular}{c|c|cccccc}
\toprule[1pt]
\multirow{2}{*}{\textbf{task name}} &\multirow{2}{*}{\textbf{model name}}& \multirow{2}{*}{\textbf{bleu}} & \multirow{2}{*}{\textbf{rouge-L}} & \multirow{2}{*}{\textbf{bertScore}} & \multicolumn{2}{c}{\textbf{RAGQuestEval}} & \multirow{2}{*}{\textbf{length}} \\
\cmidrule(lr){6-7}
& & & & & \textbf{precision} & \textbf{recall} & \\
\midrule 
\multirow{8}{*}{text continuation}  
& ChatGLM2-6B & $2.06$ & $13.35$ & $68.51$ & $20.68$ & $15.44$ & 363.3 \\
& Qwen-7B & $\textbf{7.10}$ & $15.31$ & $77.94$ & $28.06$ & $18.44$ & 159.6 \\
& Baichuan2-13B & $3.97$ & $14.21$ & $71.75$ & $28.62$ & $22.95$ & 358.4 \\
& Qwen-14B & $5.70$ & $18.48$ & $82.97$ & $27.89$ & $21.68$ & 240.1 \\
& GPT-3.5-turbo & $3.66$ & $17.78$ & $83.99$ & $26.96$ & $24.68$ & 367.6 \\
& GPT-4-0613 & $5.58$ & $\textbf{19.47}$ & $ \textbf{84.91}$ & $30.34$ & $\textbf{28.02}$ & 369.8 \\
& Qwen2-7B & 2.94	& 16.76	&83.82	&26.90	& 23.68	&  350.0 \\
& GPT-4o & 4.48	&18.85	&84.45	&\textbf{30.89}	&26.11 & 356.7 \\

\midrule

\multirow{8}{*}{summarization} & ChatGLM2-6B & $17.09$ & $28.16$ & $83.00$ & $58.94$ & $40.35$ & 228.1 \\
& Qwen-7B & $28.30$ & $30.21$ & $84.26$ & $67.62$ & $40.03$ & 240.5 \\
& Baichuan2-13B & $24.49$ & $32.49$ & $85.64$ & $65.96$ & $42.53$ & 179.5 \\
& Qwen-14B & $\textbf{32.51}$ & $33.33$ & $85.62$ & $68.94$ & $40.57$ & 139.1 \\
& GPT-3.5-turbo & $23.69$ & $33.53$ & $88.49$ & $68.06$ & $46.18$ & 205.9 \\
& GPT-4-0613 & $24.54$ & $\textbf{35.91}$ & $89.39$ & $\textbf{71.24}$ & $50.53$ & 194.6 \\
& Qwen2-7B & 14.82	&30.00	&88.60	&62.04	&45.93&	283.2  \\
& GPT-4o & 23.24&	35.40&	\textbf{89.65}	&68.28	&\textbf{50.93}	&217.7 \\

\midrule

\multirow{8}{*}{question answering} & ChatGLM2-6B & $29.11$ & $47.57$ & $79.59$ & $50.06$ & $69.35$ & 90.8 \\ 
\multirow{8}{*}{1-document} & Qwen-7B & $39.63$ & $56.71$ & $82.64$ & $51.77$ & $72.02$ & 68.8 \\
& Baichuan2-13B & $35.40$ & $53.85$ & $83.59$ & $54.35$ & $\textbf{76.92}$ & 91.3 \\
& Qwen-14B & $37.95$ & $55.13$ & $83.25$ & $53.03$ & $73.92$ & 73.8 \\
& GPT-3.5-turbo & $\textbf{39.76}$ & $\textbf{57.24}$ & $\textbf{83.81}$ & $52.67$ & $70.82$ & 73.3 \\
& GPT-4-0613 & $33.87$ & $51.42$ & $80.92$ & $53.14$ & $62.39$ & 95.9 \\
& Qwen2-7B & 23.06&	41.25	&82.10	&60.07	&72.17&	123.3 \\
& GPT-4o & 33.32&	51.78&	83.35&	\textbf{65.33}	&66.59&	74.7 \\

\midrule
\multirow{8}{*}{question answering} & ChatGLM2-6B & $15.15$ & $29.12$ & $82.30$ & $37.61$ & $51.51$ & 193.4 \\
\multirow{8}{*}{2-document} & Qwen-7B & $22.61$ & $36.07$ & $85.84$ & $42.32$ & $56.26$ & 157.6 \\
& Baichuan2-13B & $20.32$ & $35.56$ & $87.49$ & $45.01$ & $61.47$ & 208.8 \\
& Qwen-14B & $21.11$ & $34.97$ & $85.87$ & $42.23$ & $56.59$ & 151.1 \\
& GPT-3.5-turbo & $\textbf{22.75}$ & $\textbf{37.25}$ & $87.16$ & $42.93$ & $56.73$ & 149.8 \\
& GPT-4-0613 & $20.38$ & $36.08$ & $88.10$ & $\textbf{49.56}$ & $62.56$ & 223.0 \\
& Qwen2-7B & 15.26	&41.25&	82.10&	48.89&	61.41&	209.1 \\
& GPT-4o & 22.84	&36.61&	\textbf{88.38}	&44.04	&\textbf{67.44}&	124.3 \\

\midrule

\multirow{8}{*}{question answering} & ChatGLM2-6B & $14.01$ & $27.71$ & $83.42$ & $35.60$ & $45.28$ & 204.1 \\
\multirow{8}{*}{3-document} & Qwen-7B & $21.63$ & $33.42$ & $86.31$ & $39.14$ & $50.55$ & 160.6 \\
& Baichuan2-13B & $18.30$ & $33.34$ & $88.08$ & $41.35$ & $55.75$ & 227.5 \\
& Qwen-14B & $19.83$ & $33.33$ & $86.93$ & $42.01$ & $51.70$ & 161.2 \\
& GPT-3.5-turbo & $21.05$ & $35.04$ & $87.81$ & $40.32$ & $51.37$ & 156.6 \\
& GPT-4-0613 & $19.11$ & $34.58$ & $88.88$ & $\textbf{48.24}$ & $56.48$ & 235.1 \\
&  Qwen2-7B & 16.23&	32.18	&87.69	&45.72&	55.29&	207.2 \\
 & GPT-4o & \textbf{22.84}&	\textbf{35.98}	&\textbf{89.21}&	43.56	&\textbf{63.90	}&139.9 \\

\midrule

\multirow{8}{*}{hallucination} & ChatGLM2-6B & $13.51$ & $28.70$ & $71.26$ & $59.63$ & $73.02$ & 176.0 \\
\multirow{8}{*}{modification} & Qwen-7B & $22.87$ & $38.10$ & $73.52$ & $60.00$ & $73.72$ & 172.5 \\
& Baichuan2-13B & $10.56$ & $27.28$ & $68.90$ & $54.42$ & $67.47$ & 124.8 \\
& Qwen-14B & $33.78$ & $51.90$ & $79.49$ & $67.05$ & $\textbf{84.08}$ & 89.7 \\
& GPT-3.5-turbo & $32.35$ & $53.04$ & $80.49$ & $65.07$ & $80.85$ & 64.8 \\
& GPT-4-0613 & $\textbf{36.69}$ & $\textbf{55.70}$ & $\textbf{81.27}$ & $\textbf{69.18}$ & $82.06$ & 63.5 \\
& Qwen2-7B & 31.07&	52.91	&80.25	&65.48	&79.16	& 49.3 \\
& GPT-4o & 36.73&	54.79&	80.90&	63.61	&73.75	&51.9	\\

\bottomrule[1pt] 
\end{tabular}
}
\label{tab:llm}
\end{table}

\subsection{Analyzing the Impact of LLM on RAG Performance in Different Tasks}

The core of the RAG system is an LLM, which can generate accurate and fluent answers based on the user's question and the retrieved information. In this paper, we conducted experiments on several commonly used LLMs, as Table ~\ref{tab:llm} displayed.

\textbf{Text Continuation}: The experimental results show that the larger the model parameters, the better the performance. GPT-4 surpassed other large models in all tasks, demonstrating its powerful generation ability.

\textbf{Open-Domain Multi-Document Summarization}: GPT-4 also excelled in the summary generation task. It achieved higher scores than other models on the overall semantic accuracy metric, as well as the key information recall and precision metric. Moreover, the summary generated by GPT-4 was relatively concise, avoiding redundant information. GPT-4 is the most suitable model for this task.

\textbf{Question Answering}: For single-document QA, which only requires extracting relevant information from a sentence in the text, this task is relatively simple. Qwen and Baichuan2 even outperformed GPT series models. However, for multi-document QA that requires a comprehensive understanding of multiple documents, GPT-4 was far ahead of other models, showing its excellent knowledge fusion ability. The Baichuan2-13B model also performed better than GPT-3.5, indicating its potential.

\textbf{Hallucination Modification}: We found that some models generated text that was too long, introducing redundant information. The hallucination modification task only requires modifying the hallucination information, retaining other information, and not introducing irrelevant information. Therefore, ChatGLM2, Qwen-7B, and Baichuan2 did not complete this task well.

In summary, the GPT-4 model performed excellently on most tasks and evaluation metrics, proving that it is a powerful LLM. Qwen-7B and Qwen-14B models also performed well, especially in the text continuation and summary generation tasks. Baichuan2-13B model was very competitive with GPT-4 in the QA task, deserving more investigation.

\textbf{Latest LLM Evaluation}: Our dataset was constructed in December 2023. To evaluate its challenge to the latest LLMs in 2024, we experimented with two newly released models: GPT-4o(Released in May 2024) and Qwen2-7b(Released in June 2024).

The results show that GPT-4o performs similarly to its predecessor GPT-4, or with some slight improvements. In contrast, Qwen2-7b demonstrates significant improvements over its predecessor Qwen-7b in multiple tasks. These findings confirm that our benchmarks remain challenging for the latest LLMs. Additionally, it is encouraging to observe that the performance of many LLMs continues to improve with each new version.

\subsection{Suggestions for Optimizing Your RAG System}
Using the benchmark we constructed, we systematically evaluated the impact of each component in the RAG system in various application scenarios. Subsequently, we offer some suggestions for future researchers aiming to optimize the performance of the RAG system. Table \ref{tab:task_based_rag_adjustments} summarizes our recommendations

The \textbf{top-k} value is a crucial parameter for the RAG system, as it determines how many documents are retrieved for each query. Depending on the scenario, the optimal top-k value may vary. For instance, in creative content generation tasks, such as text continuation, a larger top-k value is preferable. This allows the LLMs to access more diverse and relevant knowledge, resulting in richer and more accurate content. However, this also comes with a higher computational cost. In summary tasks, a moderate top-k value can strike a balance between precision and recall of information. For scenarios that require high precision, a smaller top-k value is recommended, while for scenarios that require high recall, a larger top-k value is recommended. In single-document QA, it is still recommended to use a large top-k value, which means that the answer can be determined multiple times. In QA tasks that involve reasoning across multiple documents, a larger top-k value can help to retrieve two related and complementary articles, thus enhancing the question answering performance.

\begin{table}[t]
\centering
\caption{Recommendations for Adjusting RAG System Key Parameters Based on Different Tasks}
\resizebox{\textwidth}{!}{
\begin{tabular}{p{1.95cm}|p{2.5cm}|p{2.5cm}|p{2.5cm}|p{2.5cm}|p{2.5cm}}
\toprule[1pt]
\textbf{Scenario} & \textbf{top-k} & \textbf{Chunk Size} & \textbf{Chunk overlap} & \textbf{Retriever} & \textbf{LLM} \\
\midrule
Create: Creative Content Generation & Larger, to access diverse knowledge & Larger, to preserve article structure & Larger, to maintain semantic coherence & Dense algorithm for semantic understanding & Qwen-14B for cost-effective high-quality text \\
\midrule
Delete: Summarization & Moderate, for precision-recall balance & Smaller for more recall, larger for more precision & Larger, to maintain semantic coherence & BM25 for precise content, Dense algorithm for more recall & Qwen-14B for high-quality summaries \\
\midrule
Read: Single-document QA & Larger, for repeated determination & Moderate, for pinpointing short answers & Larger, to maintain semantic coherence &  Hybrid + rerank for enhanced performance & Baichuan2-13B for GPT-4-like performance \\
\midrule
Read: Multi-document QA & Larger, for retrieving complementary articles & Larger, for article completeness & Larger, to maintain semantic coherence  & Hybrid + rerank for enhanced performance & Baichuan2-13B for GPT-4-like performance \\
\midrule
Update: Error Correction & Smaller, for high precision tasks & Larger, to avoid breaking article structure & Smaller, error correction tasks are not sensitive to semantic coherence & BM25 for precise content generation & GPT-4 or alternatives depending on cost \\
\bottomrule[1pt]
\end{tabular}
}
\label{tab:task_based_rag_adjustments}
\end{table}

The \textbf{chunk size} is also an important factor when building the vector index for external knowledge. For creative scenarios, such as content generation, we suggest using a larger chunk size to preserve the structure of the article and avoid affecting the performance of the RAG system. For summary scenarios, a smaller chunk size can be used if more information is desired to be recalled; however, if the precision of the generated content is more important, a larger chunk size is still recommended to avoid destroying the structure of the article. In factual question answering scenarios, a smaller chunk size is beneficial for finding the answer in a short sentence. For reasoning tasks, a larger chunk size can ensure the article's completeness and enhance the reasoning ability.

The \textbf{chunk overlap} is the shared content between two adjacent chunks, and chunk overlap is key to maintaining the coherence of semantics in LLMs when dealing with long texts. Experiments show that for creative generation, summarization, and question answering scenarios, the semantic consistency between chunks is very important, so a large chunk overlap value should be maintained. However, for error correction scenarios, the semantic consistency between chunks is relatively unimportant, and a smaller chunk overlap value can be considered.

When choosing an \textbf{embedding} model, you can refer to the mteb leaderboard~\cite{muennighoff2022mteb}, which shows the performance of different embedding models on retrieval tasks. However, the actual performance of the RAG system may differ from the leaderboard, so you need to evaluate and adjust according to the specific scenario.

When choosing a \textbf{retrieval algorithm}, BM25 has the advantage of saving computational resources compared to dense retrievers, and since it is a keyword-based algorithm, it can usually retrieve very relevant documents. However, keyword-based algorithms perform poorly in capturing semantics and may miss some relevant content. Therefore, we suggest using BM25 for tasks that require precise content generation, such as hallucination modification and summarization.

However, BM25 may not be suitable for tasks that require semantic understanding, such as question answering and creative generation, and we recommend using dense algorithms based on deep learning embeddings instead.

Moreover, the hybrid algorithm that combines dense and BM25 retriever has very limited improvement on the overall quality of the generated results. However, by using a rerank model to reorder the retrieval results and then inputting them into LLMs, the performance of almost all tasks improved, especially reasoning tasks. Therefore, we suggest trying to use the hybrid algorithm + rerank retrieval mode when the conditions permit, which can achieve better performance in the RAG system.

When choosing a \textbf{large language model}, GPT-4 model is undoubtedly the most advanced model at present. However, due to the high cost of invoking GPT-4, we may need to consider some open-source alternatives. According to our experimental results, Qwen-14B model has shown similar performance to GPT-4 in the two tasks of text continuation and summary generation, and can generate high-quality creative and summarizing texts. In the QA task, Baichuan2-13B model also showed a level close to GPT-4, and can generate accurate and fluent answers. Therefore, we can choose a suitable LLM according to different tasks and cost requirements.

\section{conclusion}
In this paper, we have introduced an innovative framework (CRUD-RAG) for evaluating retrieval-augmented generation (RAG) systems that is both comprehensive and scenario-specific. Our unique categorization of text generation tasks into the CRUD—Create, Read, Update, and Delete—types provides a structured approach to assess the capabilities and limitations of RAG systems in handling a variety of textual contexts. To facilitate this evaluation, we have meticulously constructed large-scale datasets for each CRUD category, which are tailored to challenge and reflect the performance of RAG systems under different operational conditions. Through rigorous experimental comparisons, we have demonstrated that RAG systems can significantly enhance the quality of generated content by effectively incorporating information from external knowledge sources.

Our study delves into the intricate balance required in the fine-tuning process of RAG systems, highlighting the importance of optimizing the retrieval model, context length, construction of the knowledge base, and the deployment of the underlying large language model to achieve the best results. The insights provided by our findings offer a valuable roadmap for researchers and practitioners in the field, guiding them in the development and refinement of RAG systems. We believe that the methodologies and results presented in this paper will spur further exploration and innovation in the realm of RAG technologies. Our work aims to catalyze advancements in text generation applications, pushing the envelope of what is possible with the integration of retrieval mechanisms and language models. We hope that this contribution will serve as a cornerstone for future research efforts, fostering the creation of more intelligent, adaptive, and context-aware generative systems.

\bibliographystyle{ACM-Reference-Format}
\bibliography{sample-base}


\begin{thebibliography}{62}


\ifx \showCODEN    \undefined \def \showCODEN     #1{\unskip}     \fi
\ifx \showDOI      \undefined \def \showDOI       #1{#1}\fi
\ifx \showISBNx    \undefined \def \showISBNx     #1{\unskip}     \fi
\ifx \showISBNxiii \undefined \def \showISBNxiii  #1{\unskip}     \fi
\ifx \showISSN     \undefined \def \showISSN      #1{\unskip}     \fi
\ifx \showLCCN     \undefined \def \showLCCN      #1{\unskip}     \fi
\ifx \shownote     \undefined \def \shownote      #1{#1}          \fi
\ifx \showarticletitle \undefined \def \showarticletitle #1{#1}   \fi
\ifx \showURL      \undefined \def \showURL       {\relax}        \fi
\providecommand\bibfield[2]{#2}
\providecommand\bibinfo[2]{#2}
\providecommand\natexlab[1]{#1}
\providecommand\showeprint[2][]{arXiv:#2}

\bibitem[Aliannejadi et~al\mbox{.}(2019)]%
        {aliannejadi2019asking}
\bibfield{author}{\bibinfo{person}{Mohammad Aliannejadi},
  \bibinfo{person}{Hamed Zamani}, \bibinfo{person}{Fabio Crestani}, {and}
  \bibinfo{person}{W~Bruce Croft}.} \bibinfo{year}{2019}\natexlab{}.
\newblock \showarticletitle{Asking clarifying questions in open-domain
  information-seeking conversations}. In \bibinfo{booktitle}{\emph{Proceedings
  of the 42nd international acm sigir conference on research and development in
  information retrieval}}. \bibinfo{pages}{475--484}.
\newblock


\bibitem[Asai et~al\mbox{.}(2023)]%
        {DBLP:conf/acm/AsaiMZC23}
\bibfield{author}{\bibinfo{person}{Akari Asai}, \bibinfo{person}{Sewon Min},
  \bibinfo{person}{Zexuan Zhong}, {and} \bibinfo{person}{Danqi Chen}.}
  \bibinfo{year}{2023}\natexlab{}.
\newblock \showarticletitle{Retrieval-based Language Models and Applications}.
  In \bibinfo{booktitle}{\emph{Proceedings of the 61st Annual Meeting of the
  Association for Computational Linguistics: Tutorial Abstracts, {ACL} 2023,
  Toronto, Canada, July 9-14, 2023}}. \bibinfo{pages}{41--46}.
\newblock


\bibitem[B{\'e}n{\'e}dict et~al\mbox{.}(2023)]%
        {benedict2023gen}
\bibfield{author}{\bibinfo{person}{Garbiel B{\'e}n{\'e}dict},
  \bibinfo{person}{Ruqing Zhang}, {and} \bibinfo{person}{Donald Metzler}.}
  \bibinfo{year}{2023}\natexlab{}.
\newblock \showarticletitle{Gen-ir@ sigir 2023: The first workshop on
  generative information retrieval}. In \bibinfo{booktitle}{\emph{Proceedings
  of the 46th International ACM SIGIR Conference on Research and Development in
  Information Retrieval}}. \bibinfo{pages}{3460--3463}.
\newblock


\bibitem[Berntson(2023)]%
        {AAS2023}
\bibfield{author}{\bibinfo{person}{Alec Berntson}.}
  \bibinfo{year}{2023}\natexlab{}.
\newblock \bibinfo{booktitle}{\emph{Azure AI Search: Outperforming vector
  search with hybrid retrieval and ranking capabilities}}.
\newblock
\newblock
\shownote{\url{https://techcommunity.microsoft.com/t5/ai-azure-ai-services-blog/azure-ai-search-outperforming-vector-search-with-hybrid/ba-p/3929167}}.


\bibitem[Bubeck et~al\mbox{.}(2023)]%
        {bubeck2023sparks}
\bibfield{author}{\bibinfo{person}{S{\'e}bastien Bubeck},
  \bibinfo{person}{Varun Chandrasekaran}, \bibinfo{person}{Ronen Eldan},
  \bibinfo{person}{Johannes Gehrke}, \bibinfo{person}{Eric Horvitz},
  \bibinfo{person}{Ece Kamar}, \bibinfo{person}{Peter Lee},
  \bibinfo{person}{Yin~Tat Lee}, \bibinfo{person}{Yuanzhi Li},
  \bibinfo{person}{Scott Lundberg}, {et~al\mbox{.}}}
  \bibinfo{year}{2023}\natexlab{}.
\newblock \showarticletitle{Sparks of artificial general intelligence: Early
  experiments with gpt-4}.
\newblock \bibinfo{journal}{\emph{arXiv preprint arXiv:2303.12712}}
  (\bibinfo{year}{2023}).
\newblock


\bibitem[Cao et~al\mbox{.}(2020)]%
        {DBLP:conf/emnlp/CaoDWC20}
\bibfield{author}{\bibinfo{person}{Meng Cao}, \bibinfo{person}{Yue Dong},
  \bibinfo{person}{Jiapeng Wu}, {and} \bibinfo{person}{Jackie Chi~Kit Cheung}.}
  \bibinfo{year}{2020}\natexlab{}.
\newblock \showarticletitle{Factual Error Correction for Abstractive
  Summarization Models}. In \bibinfo{booktitle}{\emph{Proceedings of the 2020
  Conference on Empirical Methods in Natural Language Processing, {EMNLP} 2020,
  Online, November 16-20, 2020}}. \bibinfo{pages}{6251--6258}.
\newblock


\bibitem[Cao and Wang(2024)]%
        {cao2024verifiablegenerationsubsentencelevelfinegrained}
\bibfield{author}{\bibinfo{person}{Shuyang Cao} {and} \bibinfo{person}{Lu
  Wang}.} \bibinfo{year}{2024}\natexlab{}.
\newblock \showarticletitle{Verifiable Generation with Subsentence-Level
  Fine-Grained Citations}.
\newblock \bibinfo{journal}{\emph{arXiv preprint arXiv:2406.06125}}
  (\bibinfo{year}{2024}).
\newblock


\bibitem[Chen et~al\mbox{.}(2023b)]%
        {chen2023benchmarking}
\bibfield{author}{\bibinfo{person}{Jiawei Chen}, \bibinfo{person}{Hongyu Lin},
  \bibinfo{person}{Xianpei Han}, {and} \bibinfo{person}{Le Sun}.}
  \bibinfo{year}{2023}\natexlab{b}.
\newblock \showarticletitle{Benchmarking large language models in
  retrieval-augmented generation}.
\newblock \bibinfo{journal}{\emph{arXiv preprint arXiv:2309.01431}}
  (\bibinfo{year}{2023}).
\newblock


\bibitem[Chen et~al\mbox{.}(2023a)]%
        {chen2023hallucination}
\bibfield{author}{\bibinfo{person}{Yuyan Chen}, \bibinfo{person}{Qiang Fu},
  \bibinfo{person}{Yichen Yuan}, \bibinfo{person}{Zhihao Wen},
  \bibinfo{person}{Ge Fan}, \bibinfo{person}{Dayiheng Liu},
  \bibinfo{person}{Dongmei Zhang}, \bibinfo{person}{Zhixu Li}, {and}
  \bibinfo{person}{Yanghua Xiao}.} \bibinfo{year}{2023}\natexlab{a}.
\newblock \showarticletitle{Hallucination detection: Robustly discerning
  reliable answers in large language models}. In
  \bibinfo{booktitle}{\emph{Proceedings of the 32nd ACM International
  Conference on Information and Knowledge Management}}.
  \bibinfo{pages}{245--255}.
\newblock


\bibitem[Cheng et~al\mbox{.}(2023)]%
        {DBLP:journals/corr/abs-2305-02437}
\bibfield{author}{\bibinfo{person}{Xin Cheng}, \bibinfo{person}{Di Luo},
  \bibinfo{person}{Xiuying Chen}, \bibinfo{person}{Lemao Liu},
  \bibinfo{person}{Dongyan Zhao}, {and} \bibinfo{person}{Rui Yan}.}
  \bibinfo{year}{2023}\natexlab{}.
\newblock \showarticletitle{Lift Yourself Up: Retrieval-augmented Text
  Generation with Self Memory}.
\newblock \bibinfo{journal}{\emph{arXiv preprint arXiv:2305.02437}}
  (\bibinfo{year}{2023}).
\newblock


\bibitem[Dai et~al\mbox{.}(2023)]%
        {DBLP:conf/iclr/DaiZMLNLBGHC23}
\bibfield{author}{\bibinfo{person}{Zhuyun Dai}, \bibinfo{person}{Vincent~Y.
  Zhao}, \bibinfo{person}{Ji Ma}, \bibinfo{person}{Yi Luan},
  \bibinfo{person}{Jianmo Ni}, \bibinfo{person}{Jing Lu},
  \bibinfo{person}{Anton Bakalov}, \bibinfo{person}{Kelvin Guu},
  \bibinfo{person}{Keith~B. Hall}, {and} \bibinfo{person}{Ming{-}Wei Chang}.}
  \bibinfo{year}{2023}\natexlab{}.
\newblock \showarticletitle{Promptagator: Few-shot Dense Retrieval From 8
  Examples}. In \bibinfo{booktitle}{\emph{The Eleventh International Conference
  on Learning Representations, {ICLR} 2023, Kigali, Rwanda, May 1-5, 2023}}.
\newblock


\bibitem[Durmus et~al\mbox{.}(2020)]%
        {durmus2020feqa}
\bibfield{author}{\bibinfo{person}{Esin Durmus}, \bibinfo{person}{He He}, {and}
  \bibinfo{person}{Mona Diab}.} \bibinfo{year}{2020}\natexlab{}.
\newblock \showarticletitle{FEQA: A question answering evaluation framework for
  faithfulness assessment in abstractive summarization}.
\newblock \bibinfo{journal}{\emph{arXiv preprint arXiv:2005.03754}}
  (\bibinfo{year}{2020}).
\newblock


\bibitem[Es et~al\mbox{.}(2023)]%
        {es2023ragas}
\bibfield{author}{\bibinfo{person}{Shahul Es}, \bibinfo{person}{Jithin James},
  \bibinfo{person}{Luis Espinosa-Anke}, {and} \bibinfo{person}{Steven
  Schockaert}.} \bibinfo{year}{2023}\natexlab{}.
\newblock \showarticletitle{Ragas: Automated evaluation of retrieval augmented
  generation}.
\newblock \bibinfo{journal}{\emph{arXiv preprint arXiv:2309.15217}}
  (\bibinfo{year}{2023}).
\newblock


\bibitem[Ferrara et~al\mbox{.}(2024)]%
        {TruLens-Eval2023}
\bibfield{author}{\bibinfo{person}{Joe Ferrara}, \bibinfo{person}{Ethan-Tonic},
  {and} \bibinfo{person}{Oguzhan~Mete Ozturk}.}
  \bibinfo{year}{2024}\natexlab{}.
\newblock \bibinfo{booktitle}{\emph{The RAG Triad}}.
\newblock
\newblock
\shownote{\url{https://www.trulens.org/trulens_eval/core_concepts_rag_triad/}}.


\bibitem[Ganguly et~al\mbox{.}(2015)]%
        {ganguly2015word}
\bibfield{author}{\bibinfo{person}{Debasis Ganguly}, \bibinfo{person}{Dwaipayan
  Roy}, \bibinfo{person}{Mandar Mitra}, {and} \bibinfo{person}{Gareth~JF
  Jones}.} \bibinfo{year}{2015}\natexlab{}.
\newblock \showarticletitle{Word embedding based generalized language model for
  information retrieval}. In \bibinfo{booktitle}{\emph{Proceedings of the 38th
  international ACM SIGIR conference on research and development in information
  retrieval}}. \bibinfo{pages}{795--798}.
\newblock


\bibitem[Gao et~al\mbox{.}(2023a)]%
        {DBLP:conf/acl/GaoMLC23}
\bibfield{author}{\bibinfo{person}{Luyu Gao}, \bibinfo{person}{Xueguang Ma},
  \bibinfo{person}{Jimmy Lin}, {and} \bibinfo{person}{Jamie Callan}.}
  \bibinfo{year}{2023}\natexlab{a}.
\newblock \showarticletitle{Precise Zero-Shot Dense Retrieval without Relevance
  Labels}. In \bibinfo{booktitle}{\emph{Proceedings of the 61st Annual Meeting
  of the Association for Computational Linguistics (Volume 1: Long Papers),
  {ACL} 2023, Toronto, Canada, July 9-14, 2023}}. \bibinfo{pages}{1762--1777}.
\newblock


\bibitem[Gao et~al\mbox{.}(2023c)]%
        {DBLP:conf/emnlp/GaoYYC23}
\bibfield{author}{\bibinfo{person}{Tianyu Gao}, \bibinfo{person}{Howard Yen},
  \bibinfo{person}{Jiatong Yu}, {and} \bibinfo{person}{Danqi Chen}.}
  \bibinfo{year}{2023}\natexlab{c}.
\newblock \showarticletitle{Enabling Large Language Models to Generate Text
  with Citations}. In \bibinfo{booktitle}{\emph{Proceedings of the 2023
  Conference on Empirical Methods in Natural Language Processing, {EMNLP} 2023,
  Singapore, December 6-10, 2023}}. \bibinfo{pages}{6465--6488}.
\newblock


\bibitem[Gao et~al\mbox{.}(2023b)]%
        {DBLP:journals/corr/abs-2312-10997}
\bibfield{author}{\bibinfo{person}{Yunfan Gao}, \bibinfo{person}{Yun Xiong},
  \bibinfo{person}{Xinyu Gao}, \bibinfo{person}{Kangxiang Jia},
  \bibinfo{person}{Jinliu Pan}, \bibinfo{person}{Yuxi Bi}, \bibinfo{person}{Yi
  Dai}, \bibinfo{person}{Jiawei Sun}, \bibinfo{person}{Qianyu Guo},
  \bibinfo{person}{Meng Wang}, {and} \bibinfo{person}{Haofen Wang}.}
  \bibinfo{year}{2023}\natexlab{b}.
\newblock \showarticletitle{Retrieval-Augmented Generation for Large Language
  Models: {A} Survey}.
\newblock \bibinfo{journal}{\emph{arXiv preprint arXiv:2312.10997}}
  (\bibinfo{year}{2023}).
\newblock


\bibitem[He et~al\mbox{.}(2022)]%
        {he2022rethinking}
\bibfield{author}{\bibinfo{person}{Hangfeng He}, \bibinfo{person}{Hongming
  Zhang}, {and} \bibinfo{person}{Dan Roth}.} \bibinfo{year}{2022}\natexlab{}.
\newblock \showarticletitle{Rethinking with retrieval: Faithful large language
  model inference}.
\newblock \bibinfo{journal}{\emph{arXiv preprint arXiv:2301.00303}}
  (\bibinfo{year}{2022}).
\newblock


\bibitem[Ilin(2023)]%
        {ART2023}
\bibfield{author}{\bibinfo{person}{Ivan Ilin}.}
  \bibinfo{year}{2023}\natexlab{}.
\newblock \bibinfo{booktitle}{\emph{Advanced RAG Techniques: an Illustrated
  Overview}}.
\newblock
\newblock
\shownote{\url{https://pub.towardsai.net/advanced-rag-techniques-an-illustrated-overview-04d193d8fec6}}.


\bibitem[Izacard et~al\mbox{.}(2022)]%
        {DBLP:journals/corr/abs-2208-03299}
\bibfield{author}{\bibinfo{person}{Gautier Izacard}, \bibinfo{person}{Patrick
  S.~H. Lewis}, \bibinfo{person}{Maria Lomeli}, \bibinfo{person}{Lucas
  Hosseini}, \bibinfo{person}{Fabio Petroni}, \bibinfo{person}{Timo Schick},
  \bibinfo{person}{Jane Dwivedi{-}Yu}, \bibinfo{person}{Armand Joulin},
  \bibinfo{person}{Sebastian Riedel}, {and} \bibinfo{person}{Edouard Grave}.}
  \bibinfo{year}{2022}\natexlab{}.
\newblock \showarticletitle{Few-shot Learning with Retrieval Augmented Language
  Models}.
\newblock \bibinfo{journal}{\emph{arXiv preprint arXiv:2208.03299}}
  (\bibinfo{year}{2022}).
\newblock


\bibitem[Ji et~al\mbox{.}(2024)]%
        {DBLP:conf/aaai/JiLDN24}
\bibfield{author}{\bibinfo{person}{Bin Ji}, \bibinfo{person}{Huijun Liu},
  \bibinfo{person}{Mingzhe Du}, {and} \bibinfo{person}{See{-}Kiong Ng}.}
  \bibinfo{year}{2024}\natexlab{}.
\newblock \showarticletitle{Chain-of-Thought Improves Text Generation with
  Citations in Large Language Models}. In
  \bibinfo{booktitle}{\emph{Thirty-Eighth {AAAI} Conference on Artificial
  Intelligence, {AAAI} 2024, Thirty-Sixth Conference on Innovative Applications
  of Artificial Intelligence, {IAAI} 2024, Fourteenth Symposium on Educational
  Advances in Artificial Intelligence, {EAAI} 2014, February 20-27, 2024,
  Vancouver, Canada}}. \bibinfo{pages}{18345--18353}.
\newblock


\bibitem[Jiang et~al\mbox{.}(2023)]%
        {DBLP:conf/emnlp/JiangXGSLDYCN23}
\bibfield{author}{\bibinfo{person}{Zhengbao Jiang}, \bibinfo{person}{Frank~F.
  Xu}, \bibinfo{person}{Luyu Gao}, \bibinfo{person}{Zhiqing Sun},
  \bibinfo{person}{Qian Liu}, \bibinfo{person}{Jane Dwivedi{-}Yu},
  \bibinfo{person}{Yiming Yang}, \bibinfo{person}{Jamie Callan}, {and}
  \bibinfo{person}{Graham Neubig}.} \bibinfo{year}{2023}\natexlab{}.
\newblock \showarticletitle{Active Retrieval Augmented Generation}. In
  \bibinfo{booktitle}{\emph{Proceedings of the 2023 Conference on Empirical
  Methods in Natural Language Processing, {EMNLP} 2023, Singapore, December
  6-10, 2023}}. \bibinfo{pages}{7969--7992}.
\newblock


\bibitem[Kang et~al\mbox{.}(2023)]%
        {DBLP:journals/corr/abs-2305-18846}
\bibfield{author}{\bibinfo{person}{Minki Kang}, \bibinfo{person}{Jin~Myung
  Kwak}, \bibinfo{person}{Jinheon Baek}, {and} \bibinfo{person}{Sung~Ju
  Hwang}.} \bibinfo{year}{2023}\natexlab{}.
\newblock \showarticletitle{Knowledge Graph-Augmented Language Models for
  Knowledge-Grounded Dialogue Generation}.
\newblock \bibinfo{journal}{\emph{arXiv preprint arXiv:2305.18846}}
  (\bibinfo{year}{2023}).
\newblock


\bibitem[Kilov(1990)]%
        {kilov1990semantic}
\bibfield{author}{\bibinfo{person}{Haim Kilov}.}
  \bibinfo{year}{1990}\natexlab{}.
\newblock \showarticletitle{From semantic to object-oriented data modeling}. In
  \bibinfo{booktitle}{\emph{Systems Integration'90. Proceedings of the First
  International Conference on Systems Integration}}. IEEE,
  \bibinfo{pages}{385--393}.
\newblock


\bibitem[Kwiatkowski et~al\mbox{.}(2019)]%
        {kwiatkowski2019natural}
\bibfield{author}{\bibinfo{person}{Tom Kwiatkowski},
  \bibinfo{person}{Jennimaria Palomaki}, \bibinfo{person}{Olivia Redfield},
  \bibinfo{person}{Michael Collins}, \bibinfo{person}{Ankur Parikh},
  \bibinfo{person}{Chris Alberti}, \bibinfo{person}{Danielle Epstein},
  \bibinfo{person}{Illia Polosukhin}, \bibinfo{person}{Jacob Devlin},
  \bibinfo{person}{Kenton Lee}, {et~al\mbox{.}}}
  \bibinfo{year}{2019}\natexlab{}.
\newblock \showarticletitle{Natural questions: a benchmark for question
  answering research}.
\newblock \bibinfo{journal}{\emph{Transactions of the Association for
  Computational Linguistics}}  \bibinfo{volume}{7} (\bibinfo{year}{2019}),
  \bibinfo{pages}{453--466}.
\newblock


\bibitem[Langchain(2023)]%
        {ERABT2023}
\bibfield{author}{\bibinfo{person}{Langchain}.}
  \bibinfo{year}{2023}\natexlab{}.
\newblock \bibinfo{booktitle}{\emph{Evaluating RAG Architectures on Benchmark
  Tasks}}.
\newblock
\newblock
\shownote{\url{https://langchain-ai.github.io/langchain-benchmarks/notebooks/retrieval/comparing_techniques.html}}.


\bibitem[Lewis et~al\mbox{.}(2020)]%
        {DBLP:conf/nips/LewisPPPKGKLYR020}
\bibfield{author}{\bibinfo{person}{Patrick S.~H. Lewis}, \bibinfo{person}{Ethan
  Perez}, \bibinfo{person}{Aleksandra Piktus}, \bibinfo{person}{Fabio Petroni},
  \bibinfo{person}{Vladimir Karpukhin}, \bibinfo{person}{Naman Goyal},
  \bibinfo{person}{Heinrich K{\"{u}}ttler}, \bibinfo{person}{Mike Lewis},
  \bibinfo{person}{Wen{-}tau Yih}, \bibinfo{person}{Tim Rockt{\"{a}}schel},
  \bibinfo{person}{Sebastian Riedel}, {and} \bibinfo{person}{Douwe Kiela}.}
  \bibinfo{year}{2020}\natexlab{}.
\newblock \showarticletitle{Retrieval-Augmented Generation for
  Knowledge-Intensive {NLP} Tasks}. In \bibinfo{booktitle}{\emph{Advances in
  Neural Information Processing Systems 33: Annual Conference on Neural
  Information Processing Systems 2020, NeurIPS 2020, December 6-12, 2020,
  virtual}}.
\newblock


\bibitem[Li et~al\mbox{.}(2024)]%
        {li2024citationenhancedgenerationllmbasedchatbots}
\bibfield{author}{\bibinfo{person}{Weitao Li}, \bibinfo{person}{Junkai Li},
  \bibinfo{person}{Weizhi Ma}, {and} \bibinfo{person}{Yang Liu}.}
  \bibinfo{year}{2024}\natexlab{}.
\newblock \showarticletitle{Citation-Enhanced Generation for LLM-based
  Chatbots}.
\newblock \bibinfo{journal}{\emph{arXiv preprint arXiv:2402.16063}}
  (\bibinfo{year}{2024}).
\newblock


\bibitem[Li et~al\mbox{.}(2023)]%
        {li2023chatgpt}
\bibfield{author}{\bibinfo{person}{Xianzhi Li}, \bibinfo{person}{Samuel Chan},
  \bibinfo{person}{Xiaodan Zhu}, \bibinfo{person}{Yulong Pei},
  \bibinfo{person}{Zhiqiang Ma}, \bibinfo{person}{Xiaomo Liu}, {and}
  \bibinfo{person}{Sameena Shah}.} \bibinfo{year}{2023}\natexlab{}.
\newblock \showarticletitle{Are ChatGPT and GPT-4 General-Purpose Solvers for
  Financial Text Analytics? A Study on Several Typical Tasks}. In
  \bibinfo{booktitle}{\emph{Proceedings of the 2023 Conference on Empirical
  Methods in Natural Language Processing: Industry Track}}.
  \bibinfo{pages}{408--422}.
\newblock


\bibitem[Liang et~al\mbox{.}(2023)]%
        {liang2023uhgeval}
\bibfield{author}{\bibinfo{person}{Xun Liang}, \bibinfo{person}{Shichao Song},
  \bibinfo{person}{Simin Niu}, \bibinfo{person}{Zhiyu Li},
  \bibinfo{person}{Feiyu Xiong}, \bibinfo{person}{Bo Tang},
  \bibinfo{person}{Zhaohui Wy}, \bibinfo{person}{Dawei He},
  \bibinfo{person}{Peng Cheng}, \bibinfo{person}{Zhonghao Wang},
  {et~al\mbox{.}}} \bibinfo{year}{2023}\natexlab{}.
\newblock \showarticletitle{UHGEval: Benchmarking the Hallucination of Chinese
  Large Language Models via Unconstrained Generation}.
\newblock \bibinfo{journal}{\emph{arXiv preprint arXiv:2311.15296}}
  (\bibinfo{year}{2023}).
\newblock


\bibitem[Liu et~al\mbox{.}(2023)]%
        {DBLP:conf/emnlp/LiuZL23}
\bibfield{author}{\bibinfo{person}{Nelson~F. Liu}, \bibinfo{person}{Tianyi
  Zhang}, {and} \bibinfo{person}{Percy Liang}.}
  \bibinfo{year}{2023}\natexlab{}.
\newblock \showarticletitle{Evaluating Verifiability in Generative Search
  Engines}. In \bibinfo{booktitle}{\emph{Findings of the Association for
  Computational Linguistics: {EMNLP} 2023, Singapore, December 6-10, 2023}}.
  \bibinfo{pages}{7001--7025}.
\newblock


\bibitem[Liu et~al\mbox{.}(2024)]%
        {10.1145/3639818}
\bibfield{author}{\bibinfo{person}{Qi Liu}, \bibinfo{person}{Gang Guo},
  \bibinfo{person}{Jiaxin Mao}, \bibinfo{person}{Zhicheng Dou},
  \bibinfo{person}{Ji-Rong Wen}, \bibinfo{person}{Hao Jiang},
  \bibinfo{person}{Xinyu Zhang}, {and} \bibinfo{person}{Zhao Cao}.}
  \bibinfo{year}{2024}\natexlab{}.
\newblock \showarticletitle{An Analysis on Matching Mechanisms and Token
  Pruning for Late-interaction Models}.
\newblock \bibinfo{journal}{\emph{ACM Transactions on Information Systems
  (TOIS)}} (\bibinfo{date}{jan} \bibinfo{year}{2024}).
\newblock


\bibitem[Liu and Lapata(2019)]%
        {liu2019hierarchical}
\bibfield{author}{\bibinfo{person}{Yang Liu} {and} \bibinfo{person}{Mirella
  Lapata}.} \bibinfo{year}{2019}\natexlab{}.
\newblock \showarticletitle{Hierarchical transformers for multi-document
  summarization}.
\newblock \bibinfo{journal}{\emph{arXiv preprint arXiv:1905.13164}}
  (\bibinfo{year}{2019}).
\newblock


\bibitem[Lyu et~al\mbox{.}(2022)]%
        {lyu2022faithful}
\bibfield{author}{\bibinfo{person}{Yuanjie Lyu}, \bibinfo{person}{Chen Zhu},
  \bibinfo{person}{Tong Xu}, \bibinfo{person}{Zikai Yin}, {and}
  \bibinfo{person}{Enhong Chen}.} \bibinfo{year}{2022}\natexlab{}.
\newblock \showarticletitle{Faithful Abstractive Summarization via Fact-aware
  Consistency-constrained Transformer}. In
  \bibinfo{booktitle}{\emph{Proceedings of the 31st ACM International
  Conference on Information \& Knowledge Management}}.
  \bibinfo{pages}{1410--1419}.
\newblock


\bibitem[Ma et~al\mbox{.}(2023b)]%
        {DBLP:journals/corr/abs-2305-14283}
\bibfield{author}{\bibinfo{person}{Xinbei Ma}, \bibinfo{person}{Yeyun Gong},
  \bibinfo{person}{Pengcheng He}, \bibinfo{person}{Hai Zhao}, {and}
  \bibinfo{person}{Nan Duan}.} \bibinfo{year}{2023}\natexlab{b}.
\newblock \showarticletitle{Query Rewriting for Retrieval-Augmented Large
  Language Models}.
\newblock \bibinfo{journal}{\emph{arXiv preprint arXiv:2305.14283}}
  (\bibinfo{year}{2023}).
\newblock


\bibitem[Ma et~al\mbox{.}(2023a)]%
        {DBLP:conf/emnlp/Ma0HS23}
\bibfield{author}{\bibinfo{person}{Yubo Ma}, \bibinfo{person}{Yixin Cao},
  \bibinfo{person}{Yong Hong}, {and} \bibinfo{person}{Aixin Sun}.}
  \bibinfo{year}{2023}\natexlab{a}.
\newblock \showarticletitle{Large Language Model Is Not a Good Few-shot
  Information Extractor, but a Good Reranker for Hard Samples!}. In
  \bibinfo{booktitle}{\emph{Findings of the Association for Computational
  Linguistics: {EMNLP} 2023, Singapore, December 6-10, 2023}}.
  \bibinfo{pages}{10572--10601}.
\newblock


\bibitem[Muennighoff et~al\mbox{.}(2022)]%
        {muennighoff2022mteb}
\bibfield{author}{\bibinfo{person}{Niklas Muennighoff},
  \bibinfo{person}{Nouamane Tazi}, \bibinfo{person}{Lo{\"\i}c Magne}, {and}
  \bibinfo{person}{Nils Reimers}.} \bibinfo{year}{2022}\natexlab{}.
\newblock \showarticletitle{MTEB: Massive Text Embedding Benchmark}.
\newblock \bibinfo{journal}{\emph{arXiv preprint arXiv:2210.07316}}
  (\bibinfo{year}{2022}).
\newblock
\urldef\tempurl%
\url{https://doi.org/10.48550/ARXIV.2210.07316}
\showDOI{\tempurl}


\bibitem[Oberst(2023)]%
        {HELR2023}
\bibfield{author}{\bibinfo{person}{Darren Oberst}.}
  \bibinfo{year}{2023}\natexlab{}.
\newblock \bibinfo{booktitle}{\emph{How to Evaluate LLMs for RAG?}}
\newblock
\newblock
\shownote{\url{https://medium.com/@darrenoberst/how-accurate-is-rag-8f0706281fd9}}.


\bibitem[Petroni et~al\mbox{.}(2020)]%
        {petroni2020kilt}
\bibfield{author}{\bibinfo{person}{Fabio Petroni}, \bibinfo{person}{Aleksandra
  Piktus}, \bibinfo{person}{Angela Fan}, \bibinfo{person}{Patrick Lewis},
  \bibinfo{person}{Majid Yazdani}, \bibinfo{person}{Nicola De~Cao},
  \bibinfo{person}{James Thorne}, \bibinfo{person}{Yacine Jernite},
  \bibinfo{person}{Vladimir Karpukhin}, \bibinfo{person}{Jean Maillard},
  {et~al\mbox{.}}} \bibinfo{year}{2020}\natexlab{}.
\newblock \showarticletitle{KILT: a benchmark for knowledge intensive language
  tasks}.
\newblock \bibinfo{journal}{\emph{arXiv preprint arXiv:2009.02252}}
  (\bibinfo{year}{2020}).
\newblock


\bibitem[Qu et~al\mbox{.}(2020)]%
        {qu2020open}
\bibfield{author}{\bibinfo{person}{Chen Qu}, \bibinfo{person}{Liu Yang},
  \bibinfo{person}{Cen Chen}, \bibinfo{person}{Minghui Qiu},
  \bibinfo{person}{W~Bruce Croft}, {and} \bibinfo{person}{Mohit Iyyer}.}
  \bibinfo{year}{2020}\natexlab{}.
\newblock \showarticletitle{Open-retrieval conversational question answering}.
  In \bibinfo{booktitle}{\emph{Proceedings of the 43rd International ACM SIGIR
  conference on research and development in Information Retrieval}}.
  \bibinfo{pages}{539--548}.
\newblock


\bibitem[Rashkin et~al\mbox{.}(2023)]%
        {10.1162/coli_a_00486}
\bibfield{author}{\bibinfo{person}{Hannah Rashkin}, \bibinfo{person}{Vitaly
  Nikolaev}, \bibinfo{person}{Matthew Lamm}, \bibinfo{person}{Lora Aroyo},
  \bibinfo{person}{Michael Collins}, \bibinfo{person}{Dipanjan Das},
  \bibinfo{person}{Slav Petrov}, \bibinfo{person}{Gaurav~Singh Tomar},
  \bibinfo{person}{Iulia Turc}, {and} \bibinfo{person}{David Reitter}.}
  \bibinfo{year}{2023}\natexlab{}.
\newblock \showarticletitle{{Measuring Attribution in Natural Language
  Generation Models}}.
\newblock \bibinfo{journal}{\emph{Computational Linguistics}}
  \bibinfo{volume}{49}, \bibinfo{number}{4} (\bibinfo{year}{2023}),
  \bibinfo{pages}{777--840}.
\newblock


\bibitem[Raunak et~al\mbox{.}(2021)]%
        {DBLP:conf/naacl/RaunakMJ21}
\bibfield{author}{\bibinfo{person}{Vikas Raunak}, \bibinfo{person}{Arul
  Menezes}, {and} \bibinfo{person}{Marcin Junczys{-}Dowmunt}.}
  \bibinfo{year}{2021}\natexlab{}.
\newblock \showarticletitle{The Curious Case of Hallucinations in Neural
  Machine Translation}. In \bibinfo{booktitle}{\emph{Proceedings of the 2021
  Conference of the North American Chapter of the Association for Computational
  Linguistics: Human Language Technologies, {NAACL-HLT} 2021, Online, June
  6-11, 2021}}. \bibinfo{pages}{1172--1183}.
\newblock


\bibitem[Saad-Falcon et~al\mbox{.}(2023)]%
        {saad2023ares}
\bibfield{author}{\bibinfo{person}{Jon Saad-Falcon}, \bibinfo{person}{Omar
  Khattab}, \bibinfo{person}{Christopher Potts}, {and} \bibinfo{person}{Matei
  Zaharia}.} \bibinfo{year}{2023}\natexlab{}.
\newblock \showarticletitle{Ares: An automated evaluation framework for
  retrieval-augmented generation systems}.
\newblock \bibinfo{journal}{\emph{arXiv preprint arXiv:2311.09476}}
  (\bibinfo{year}{2023}).
\newblock


\bibitem[Scialom et~al\mbox{.}(2021)]%
        {scialom2021questeval}
\bibfield{author}{\bibinfo{person}{Thomas Scialom},
  \bibinfo{person}{Paul-Alexis Dray}, \bibinfo{person}{Patrick Gallinari},
  \bibinfo{person}{Sylvain Lamprier}, \bibinfo{person}{Benjamin Piwowarski},
  \bibinfo{person}{Jacopo Staiano}, {and} \bibinfo{person}{Alex Wang}.}
  \bibinfo{year}{2021}\natexlab{}.
\newblock \showarticletitle{Questeval: Summarization asks for fact-based
  evaluation}.
\newblock \bibinfo{journal}{\emph{arXiv preprint arXiv:2103.12693}}
  (\bibinfo{year}{2021}).
\newblock


\bibitem[Shen et~al\mbox{.}(2023)]%
        {shen2023chatgpt}
\bibfield{author}{\bibinfo{person}{Xinyue Shen}, \bibinfo{person}{Zeyuan Chen},
  \bibinfo{person}{Michael Backes}, {and} \bibinfo{person}{Yang Zhang}.}
  \bibinfo{year}{2023}\natexlab{}.
\newblock \showarticletitle{In chatgpt we trust? measuring and characterizing
  the reliability of chatgpt}.
\newblock \bibinfo{journal}{\emph{arXiv preprint arXiv:2304.08979}}
  (\bibinfo{year}{2023}).
\newblock


\bibitem[Shi et~al\mbox{.}(2023)]%
        {DBLP:journals/corr/abs-2301-12652}
\bibfield{author}{\bibinfo{person}{Weijia Shi}, \bibinfo{person}{Sewon Min},
  \bibinfo{person}{Michihiro Yasunaga}, \bibinfo{person}{Minjoon Seo},
  \bibinfo{person}{Rich James}, \bibinfo{person}{Mike Lewis},
  \bibinfo{person}{Luke Zettlemoyer}, {and} \bibinfo{person}{Wen{-}tau Yih}.}
  \bibinfo{year}{2023}\natexlab{}.
\newblock \showarticletitle{{REPLUG:} Retrieval-Augmented Black-Box Language
  Models}.
\newblock \bibinfo{journal}{\emph{arXiv preprint arXiv:2301.12652}}
  (\bibinfo{year}{2023}).
\newblock


\bibitem[Truica et~al\mbox{.}(2015)]%
        {truica2015performance}
\bibfield{author}{\bibinfo{person}{Ciprian-Octavian Truica},
  \bibinfo{person}{Florin Radulescu}, \bibinfo{person}{Alexandru Boicea}, {and}
  \bibinfo{person}{Ion Bucur}.} \bibinfo{year}{2015}\natexlab{}.
\newblock \showarticletitle{Performance evaluation for CRUD operations in
  asynchronously replicated document oriented database}. In
  \bibinfo{booktitle}{\emph{2015 20th International Conference on Control
  Systems and Computer Science}}. IEEE, \bibinfo{pages}{191--196}.
\newblock


\bibitem[Wang et~al\mbox{.}(2020)]%
        {wang2020asking}
\bibfield{author}{\bibinfo{person}{Alex Wang}, \bibinfo{person}{Kyunghyun Cho},
  {and} \bibinfo{person}{Mike Lewis}.} \bibinfo{year}{2020}\natexlab{}.
\newblock \showarticletitle{Asking and answering questions to evaluate the
  factual consistency of summaries}.
\newblock \bibinfo{journal}{\emph{arXiv preprint arXiv:2004.04228}}
  (\bibinfo{year}{2020}).
\newblock


\bibitem[Wang et~al\mbox{.}(2023)]%
        {DBLP:conf/emnlp/WangYW23}
\bibfield{author}{\bibinfo{person}{Liang Wang}, \bibinfo{person}{Nan Yang},
  {and} \bibinfo{person}{Furu Wei}.} \bibinfo{year}{2023}\natexlab{}.
\newblock \showarticletitle{Query2doc: Query Expansion with Large Language
  Models}. In \bibinfo{booktitle}{\emph{Proceedings of the 2023 Conference on
  Empirical Methods in Natural Language Processing, {EMNLP} 2023, Singapore,
  December 6-10, 2023}}. \bibinfo{pages}{9414--9423}.
\newblock


\bibitem[Wei et~al\mbox{.}(2022)]%
        {wei2022chain}
\bibfield{author}{\bibinfo{person}{Jason Wei}, \bibinfo{person}{Xuezhi Wang},
  \bibinfo{person}{Dale Schuurmans}, \bibinfo{person}{Maarten Bosma},
  \bibinfo{person}{Fei Xia}, \bibinfo{person}{Ed Chi}, \bibinfo{person}{Quoc~V
  Le}, \bibinfo{person}{Denny Zhou}, {et~al\mbox{.}}}
  \bibinfo{year}{2022}\natexlab{}.
\newblock \showarticletitle{Chain-of-thought prompting elicits reasoning in
  large language models}.
\newblock \bibinfo{journal}{\emph{Advances in Neural Information Processing
  Systems}}  \bibinfo{volume}{35} (\bibinfo{year}{2022}),
  \bibinfo{pages}{24824--24837}.
\newblock


\bibitem[Wen et~al\mbox{.}(2023)]%
        {DBLP:journals/corr/abs-2308-09729}
\bibfield{author}{\bibinfo{person}{Yilin Wen}, \bibinfo{person}{Zifeng Wang},
  {and} \bibinfo{person}{Jimeng Sun}.} \bibinfo{year}{2023}\natexlab{}.
\newblock \showarticletitle{MindMap: Knowledge Graph Prompting Sparks Graph of
  Thoughts in Large Language Models}.
\newblock \bibinfo{journal}{\emph{arXiv preprint arXiv:2308.09729}}
  (\bibinfo{year}{2023}).
\newblock


\bibitem[Xu et~al\mbox{.}(2023)]%
        {DBLP:journals/corr/abs-2310-04408}
\bibfield{author}{\bibinfo{person}{Fangyuan Xu}, \bibinfo{person}{Weijia Shi},
  {and} \bibinfo{person}{Eunsol Choi}.} \bibinfo{year}{2023}\natexlab{}.
\newblock \showarticletitle{{RECOMP:} Improving Retrieval-Augmented LMs with
  Compression and Selective Augmentation}.
\newblock \bibinfo{journal}{\emph{arXiv preprint arXiv:2310.04408}}
  (\bibinfo{year}{2023}).
\newblock


\bibitem[Xu et~al\mbox{.}(2024)]%
        {xu2024aliiceevaluatingpositionalfinegrained}
\bibfield{author}{\bibinfo{person}{Yilong Xu}, \bibinfo{person}{Jinhua Gao},
  \bibinfo{person}{Xiaoming Yu}, \bibinfo{person}{Baolong Bi},
  \bibinfo{person}{Huawei Shen}, {and} \bibinfo{person}{Xueqi Cheng}.}
  \bibinfo{year}{2024}\natexlab{}.
\newblock \showarticletitle{ALiiCE: Evaluating Positional Fine-grained Citation
  Generation}.
\newblock \bibinfo{journal}{\emph{arXiv preprint arXiv:2406.13375}}
  (\bibinfo{year}{2024}).
\newblock


\bibitem[Yang et~al\mbox{.}(2023)]%
        {DBLP:conf/emnlp/YangLZW0L023}
\bibfield{author}{\bibinfo{person}{Haoyan Yang}, \bibinfo{person}{Zhitao Li},
  \bibinfo{person}{Yong Zhang}, \bibinfo{person}{Jianzong Wang},
  \bibinfo{person}{Ning Cheng}, \bibinfo{person}{Ming Li}, {and}
  \bibinfo{person}{Jing Xiao}.} \bibinfo{year}{2023}\natexlab{}.
\newblock \showarticletitle{{PRCA:} Fitting Black-Box Large Language Models for
  Retrieval Question Answering via Pluggable Reward-Driven Contextual Adapter}.
  In \bibinfo{booktitle}{\emph{Proceedings of the 2023 Conference on Empirical
  Methods in Natural Language Processing, {EMNLP} 2023, Singapore, December
  6-10, 2023}}. \bibinfo{publisher}{Association for Computational Linguistics},
  \bibinfo{pages}{5364--5375}.
\newblock


\bibitem[Ye et~al\mbox{.}(2020)]%
        {DBLP:conf/emnlp/YeLDLLSL20}
\bibfield{author}{\bibinfo{person}{Deming Ye}, \bibinfo{person}{Yankai Lin},
  \bibinfo{person}{Jiaju Du}, \bibinfo{person}{Zhenghao Liu},
  \bibinfo{person}{Peng Li}, \bibinfo{person}{Maosong Sun}, {and}
  \bibinfo{person}{Zhiyuan Liu}.} \bibinfo{year}{2020}\natexlab{}.
\newblock \showarticletitle{Coreferential Reasoning Learning for Language
  Representation}. In \bibinfo{booktitle}{\emph{Proceedings of the 2020
  Conference on Empirical Methods in Natural Language Processing, {EMNLP} 2020,
  Online, November 16-20, 2020}}. \bibinfo{pages}{7170--7186}.
\newblock


\bibitem[Yu et~al\mbox{.}(2020)]%
        {yu2020few}
\bibfield{author}{\bibinfo{person}{Shi Yu}, \bibinfo{person}{Jiahua Liu},
  \bibinfo{person}{Jingqin Yang}, \bibinfo{person}{Chenyan Xiong},
  \bibinfo{person}{Paul Bennett}, \bibinfo{person}{Jianfeng Gao}, {and}
  \bibinfo{person}{Zhiyuan Liu}.} \bibinfo{year}{2020}\natexlab{}.
\newblock \showarticletitle{Few-shot generative conversational query
  rewriting}. In \bibinfo{booktitle}{\emph{Proceedings of the 43rd
  International ACM SIGIR conference on research and development in Information
  Retrieval}}. \bibinfo{pages}{1933--1936}.
\newblock


\bibitem[Yu et~al\mbox{.}(2023)]%
        {DBLP:journals/corr/abs-2311-09210}
\bibfield{author}{\bibinfo{person}{Wenhao Yu}, \bibinfo{person}{Hongming
  Zhang}, \bibinfo{person}{Xiaoman Pan}, \bibinfo{person}{Kaixin Ma},
  \bibinfo{person}{Hongwei Wang}, {and} \bibinfo{person}{Dong Yu}.}
  \bibinfo{year}{2023}\natexlab{}.
\newblock \showarticletitle{Chain-of-Note: Enhancing Robustness in
  Retrieval-Augmented Language Models}.
\newblock \bibinfo{journal}{\emph{arXiv preprint arXiv:2311.09210}}
  (\bibinfo{year}{2023}).
\newblock


\bibitem[Zhai and Lafferty(2004)]%
        {zhai2004study}
\bibfield{author}{\bibinfo{person}{Chengxiang Zhai} {and} \bibinfo{person}{John
  Lafferty}.} \bibinfo{year}{2004}\natexlab{}.
\newblock \showarticletitle{A study of smoothing methods for language models
  applied to information retrieval}.
\newblock \bibinfo{journal}{\emph{ACM Transactions on Information Systems
  (TOIS)}} \bibinfo{volume}{22}, \bibinfo{number}{2} (\bibinfo{year}{2004}),
  \bibinfo{pages}{179--214}.
\newblock


\bibitem[Zhang et~al\mbox{.}(2023)]%
        {DBLP:journals/corr/abs-2310-07554}
\bibfield{author}{\bibinfo{person}{Peitian Zhang}, \bibinfo{person}{Shitao
  Xiao}, \bibinfo{person}{Zheng Liu}, \bibinfo{person}{Zhicheng Dou}, {and}
  \bibinfo{person}{Jian{-}Yun Nie}.} \bibinfo{year}{2023}\natexlab{}.
\newblock \showarticletitle{Retrieve Anything To Augment Large Language
  Models}.
\newblock \bibinfo{journal}{\emph{arXiv preprint arXiv:2310.07554}}
  (\bibinfo{year}{2023}).
\newblock


\bibitem[Zhang et~al\mbox{.}(2024)]%
        {zhang2024finegrainedcitationevaluationgenerated}
\bibfield{author}{\bibinfo{person}{Weijia Zhang}, \bibinfo{person}{Mohammad
  Aliannejadi}, \bibinfo{person}{Yifei Yuan}, \bibinfo{person}{Jiahuan Pei},
  \bibinfo{person}{Jia-Hong Huang}, {and} \bibinfo{person}{Evangelos
  Kanoulas}.} \bibinfo{year}{2024}\natexlab{}.
\newblock \showarticletitle{Towards Fine-Grained Citation Evaluation in
  Generated Text: A Comparative Analysis of Faithfulness Metrics}.
\newblock \bibinfo{journal}{\emph{arXiv preprint arXiv:2406.15264}}
  (\bibinfo{year}{2024}).
\newblock


\bibitem[Zuccon et~al\mbox{.}(2023)]%
        {zuccon2023chatgpt}
\bibfield{author}{\bibinfo{person}{Guido Zuccon}, \bibinfo{person}{Bevan
  Koopman}, {and} \bibinfo{person}{Razia Shaik}.}
  \bibinfo{year}{2023}\natexlab{}.
\newblock \showarticletitle{Chatgpt hallucinates when attributing answers}. In
  \bibinfo{booktitle}{\emph{Proceedings of the Annual International ACM SIGIR
  Conference on Research and Development in Information Retrieval in the Asia
  Pacific Region}}. \bibinfo{pages}{46--51}.
\newblock


\end{thebibliography}










\end{document}